\documentclass[11pt]{article}

\usepackage[preprint]{acl}

\usepackage{times}
\usepackage{latexsym}

\usepackage[T1]{fontenc}

\usepackage[utf8]{inputenc}

\usepackage{microtype}

\usepackage{inconsolata}

\usepackage{graphicx}

\usepackage{booktabs}
\usepackage{multirow}
\usepackage{multicol}
\usepackage{makecell}
\usepackage{array}
\usepackage{CJKutf8}

\usepackage[most]{tcolorbox}
\usepackage{tabularx} 
\usepackage{ragged2e} 
\definecolor{ufv2_orange}{HTML}{faa755} 
\definecolor{ufv2_green}{HTML}{a8e6cf}
\definecolor{ufv2_purple}{HTML}{d3b5e5}

\title{\raisebox{-0.08cm}{\includegraphics[width=0.70cm]{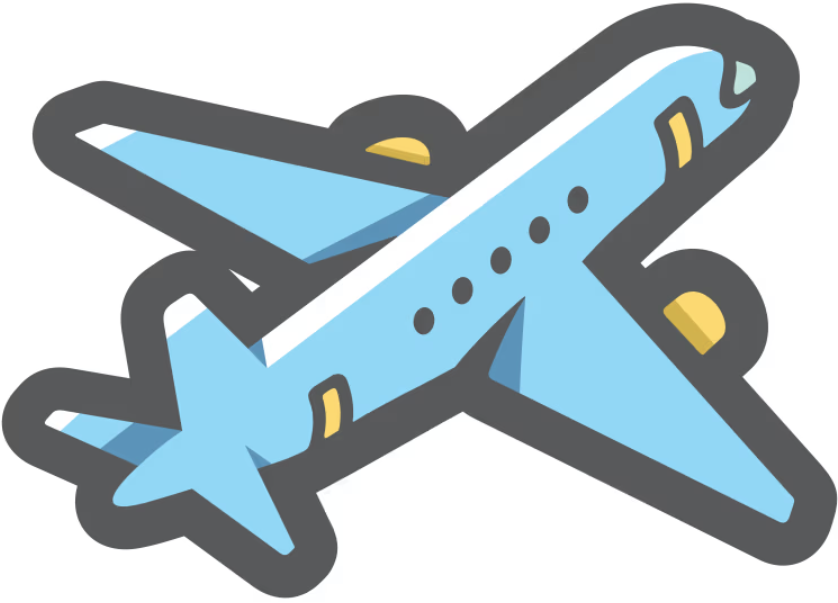}} \hspace{-0.1cm} LexPilot: Benchmarking the Proactive Interaction Ability of Legal LLMs with Laypeople in Chinese Civil Case Consultation}

\author{
  \textbf{Wenbin Zhang\textsuperscript{1}},
  \textbf{ShangZhan Li\textsuperscript{1}},
  \textbf{Bingxiang He\textsuperscript{2}},
  \textbf{Ye He\textsuperscript{1}},
  \textbf{Yuxin Zhou\textsuperscript{1}},
  \textbf{Muzhen Cai\textsuperscript{1}} \\
  \textbf{Qi Shi\textsuperscript{2}},
  \textbf{Wanxiang Che\textsuperscript{1}} \\
  \textsuperscript{1}Harbin Institute of Technology, \quad
  \textsuperscript{2}Tsinghua University
}

\begin{document}
\maketitle

\begingroup
\renewcommand{\thefootnote}{}
\footnotetext{\textit{Preprint}. Code and benchmark are publicly available at \url{https://huggingface.co/datasets/Medivvv/LexPilot}.}
\endgroup

\begin{abstract}

Large Language Models (LLMs) have greatly assisted legal professionals. However, our empirical studies reveal that, when faced with underspecified consultations from non-experts, current LLMs often produce irresponsible responses with limited utility despite lacking sufficient case details for definitive conclusions. To better assist laypeople, legal LLMs must identify focal points of dispute and proactively elicit crucial context from users. Unfortunately, existing data resources often focus on interaction and response quality, offering limited insight into how proactive interaction affects LLMs' downstream claim-level judgment prediction in actual civil cases. To address this gap, we introduce \textbf{LexPilot}, a benchmark linking proactive legal interaction to claim-level verdict prediction using authentic Chinese civil judgments. Constructed from 432 real-world judgments across 36 common causes of action, LexPilot comprises expert-reviewed pairs of complete and underspecified consultation histories with corresponding court verdicts. Experiments reveal that current LLMs struggle with proactive legal interactions in underspecified consultations, with downstream judgment prediction substantially underperforming that in the fully specified consultation setup (-23.84\% on average). Further analysis identifies critical bottlenecks hindering these capabilities. Our work highlights the urgent need for more user-centered legal AI to better serve the public.
\end{abstract}
\section{Introduction}

\begin{figure}[t]
  \centering
  \includegraphics[width=0.95\columnwidth]{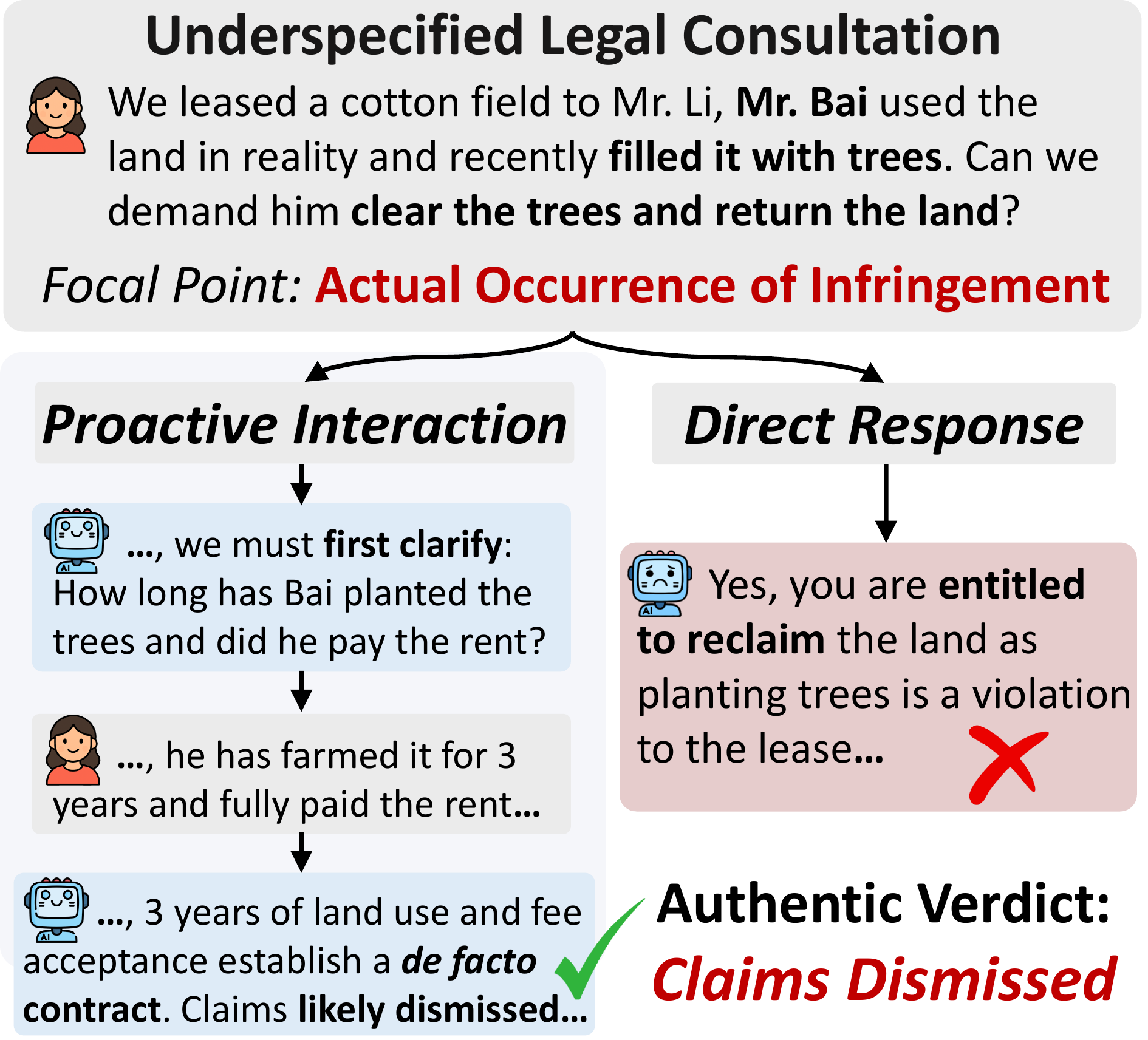}
  \caption{LLMs deliver more responsible judgment predictions via proactive interaction than direct responses.}
  \label{fig:fig1}
  \vspace{-15pt}
\end{figure}

Large Language Models (LLMs) have greatly facilitated the work of legal professionals through their outstanding performance across a range of legal tasks, such as Similar Case Retrieval~\cite{xiao2019cail2019, gao-etal-2024-enhancing-legal}, Legal Judgment Prediction~\cite{xiao2018cail2018largescalelegaldataset,nigam-etal-2024-rethinking} and Legal Event Detection~\cite{yao-etal-2022-leven}. While these works excel in professional settings where the input case descriptions are typically formal and complete, many non-experts lack adequate access to justice and exhibit strong willingness to take advantage of LLMs for accessible legal assistance~\cite{Seabrooke2024ASO, hoebenkuil2025llmscreatelegallyrelevant}. 

However, constrained by a lack of in-depth legal understanding, laypeople tend to omit crucial case details in their actual consultations, leading to ambiguous case facts surrounding the focal point of dispute~\cite{Zhang2024CitaLawEL}. Such underspecified consultations are \textit{insufficient} for legal LLMs to provide definitive conclusions; yet, as our preliminary studies will reveal, current LLMs often default to irresponsible responses with limited utility. These unreliable suggestions pose a potentially misleading risk to laypeople~\cite{10.1145/3630106.3659048, 10.1145/3706598.3713470}, calling for more reliable legal LLMs to better serve the public.

To mitigate this risk, legal LLMs should refrain from providing direct responses when the consultation history lacks essential information. Instead, they must first assess whether the available case facts are sufficient. If not, as illustrated in Figure~\ref{fig:fig1}, they should identify and proactively elicit missing details relevant to the focal points of dispute before reaching the final conclusion. This paradigm aligns with recent work on proactive interaction in LLMs~\cite{zhang-etal-2024-clamber, qian-etal-2024-tell}, motivating our evaluation of this capability in legal consultations to better assist laypeople.

Nevertheless, existing work on proactive legal interaction often evaluates interaction quality~\cite{yuan2025lecodebenchmarkdatasetinteractive}, missing element detection~\cite{insufficiencybench} and legal reasoning abilities~\cite{dlawbench}, offering limited insight into how proactive clarification of essential case details affects LLMs' downstream judgment prediction performance in real-world civil cases. To address this gap, we present \textbf{LexPilot}, a benchmark linking proactive legal interaction to claim-level judgment prediction using authentic Chinese civil judgments as ground truth.
Using 432 Chinese civil judgments spanning 36 causes of action, we construct an expert-reviewed pair of fully specified and underspecified consultation histories for each case. LexPilot decomposes proactive legal interaction into three sequential tasks: Legal Completeness Assessment (\textbf{LCA}), Detail Completion Guidance (\textbf{DCG}), and Legal Judgment Prediction (\textbf{LJP}), tracing the process from omission detection through fact elicitation to verdict prediction.

Following benchmark construction, we propose \textbf{LexAwN}, a legal-domain adaptation of the proactive interaction method AwN~\cite{wang-etal-2025-learning}, as the baseline approach to assess a wide range of general and legal LLMs. 
Results on LexPilot reveal substantial limitations in current LLMs' omission detection and proactive guidance abilities. Although proactive interaction generally improves judgment prediction for underspecified consultations, models still exhibit an average relative performance degradation of 23.84\% compared with complete consultations. Further analysis reveals variation in proactive interaction performance across case complexity levels and types of omitted details, highlighting challenges in eliciting the information needed for legal judgment prediction.

In summary, our contributions are threefold:
\begin{itemize}
\item We empirically reveal the occurrence of underspecified legal consultations from laypeople, highlighting that proactive interactions are essential for legal LLMs in this scenario.

\item We present \textbf{LexPilot}, a benchmark derived from authentic civil judgments to evaluate LLMs' proactive legal interaction ability and its impact on claim-level judgment prediction.

\item Experiments reveal current LLMs' critical deficiencies in proactive legal interaction, underscoring the need to advance user-centered legal AI to better serve the public.

\end{itemize}

\section{Related Work}
We review work on Legal Judgment Prediction and Proactive Interaction in LLMs, both of which are central to LexPilot's evaluation.

\subsection{Legal Judgment Prediction}
Legal Judgment Prediction (LJP) predicts case outcomes from case fact descriptions~\cite{cui2023survey}. Following CAIL-2018~\cite{xiao2018cail2018largescalelegaldataset}, benchmarks have covered diverse legal scenarios~\cite{zhong-etal-2018-legal, wu-etal-2020-de, zhong2020iteratively, 10.1145/3404835.3462945, 10.1145/3626772.3657717, zhang-etal-2025-beyond-guilt}. Prior work has framed LJP as a charge classification task~\cite{de_Arriba_P_rez_2022, huang-etal-2024-cmdl}, enriched case representations through event extraction and summarization~\cite{feng-etal-2022-legal, 10.1007/s10506-023-09349-8}, and adopted prompting~\cite{Trautmann2022LegalPE}, multi-agent systems~\cite{chen-etal-2025-debate}, and precedent retrieval~\cite{t-y-s-s-etal-2024-incorporating}. However, these methods generally assume professionally written case descriptions and thus overlook underspecified consultations from laypeople.

\subsection{Proactive Interaction in LLMs}
Aligning LLMs with general preferences~\cite{ouyang2022traininglanguagemodelsfollow,rafailov2023direct,he2025air,FIR} cannot fully resolve ambiguity in user interactions, motivating proactive interaction to elicit users' true intentions~\cite{qian-etal-2024-tell}. CLAMBER~\cite{zhang-etal-2024-clamber} evaluates this ability in traditional NLP tasks. Subsequent benchmarks extend proactive interaction to tool invocation~\cite{zhang-etal-2025-asktoact, wang-etal-2025-learning}, personalization~\cite{coldchat}, real-world planning~\cite{zhang-etal-2024-ask}, multi-turn dialogue~\cite{liu2025proactiveevalunifiedevaluationframework}, autonomous task initiation~\cite{lu2025proactive}, user-centric agents~\cite{qian2025userbenchinteractivegymenvironment}, and interactive medical diagnosis~\cite{liu-etal-2025-interactive}.
For proactive legal interaction, existing work targets document navigation~\cite{chouhan2025answersguidanceproactivedialogue}, omission detection~\cite{insufficiencybench}, and fact collection~\cite{yuan2025lecodebenchmarkdatasetinteractive,bystronski2026factual,wang2026rps}. Downstream tasks include legal advice~\cite{yao2025intelligentlegalassistantinteractive,yang2026collt}, court-view generation~\cite{wu-etal-2024-knowledge}, and legal intake classification~\cite{steenhuis2026wednesdays}. MASER~\cite{shengbinyue-etal-2025-multi} uses multi-agent simulation to train and evaluate interactive complaint drafting. InfoGatherer~\cite{taranukhin2026infogatherer} evaluates legal exam-derived tasks. DLawBench~\cite{dlawbench} uses court-grounded rubrics to assess fact elicitation and legal reasoning. J1-ENVS~\cite{jia-etal-2026-ready} evaluates judgment quality in simulated court proceedings. LexPilot complements these efforts through paired complete and underspecified consultations to compare interaction-based and full-information verdict prediction for the same real-world civil cases.

\section{Preliminary Studies}
\label{sec:Preliminaries}

To motivate this research, we present two pilot studies on laypeople's omission of key details and LLMs' responses to underspecified consultations.

\subsection{Essential Detail Omission in Consultations}
\label{sec:Preliminary_Studies}

To examine laypeople's omission of essential details, we conducted a controlled consultation-writing study with 16 undergraduates: 8 law majors and 8 without legal backgrounds. For each civil judgment, participants viewed the plaintiff's claims and case facts from a real case for two minutes to form an initial impression while limiting rote memorization. The materials were then hidden, and participants wrote a consultation to a legal assistant solely from memory, describing the case in detail from the plaintiff's perspective and asking whether their claims would be supported. This setup tests which facts participants retain and convey when formulating consultations.

Despite similar consultation lengths (174.2 average tokens for law majors versus 157.6 for laypeople), laypeople covered only 68.1\% of expert-annotated (detailed in Appendix~\ref{sec:human_anno_detail}) key details related to the focal points of dispute, compared to 86.6\% for law majors. This gap suggests that legal knowledge helps participants prioritize crucial facts when translating their recollections into consultations. The omissions in this controlled setting motivate legal LLMs to proactively elicit essential context from non-experts before offering advice. Appendix~\ref{sec:appendix_prelim_1_details} provides case selection, annotation criteria, and results for this pilot study.

\subsection{Tendency Towards Irresponsible Response}
\label{sec:Preliminary_Studies_Models}

Having revealed the prevalence of underspecified legal consultations among laypeople, we then analyze how LLMs respond to them. Specifically, we evaluated three advanced LLMs on their ability to handle missing essential details. Our results indicate a concerning tendency towards irresponsible responses: despite lacking crucial information necessary for a definitive prediction, models rarely inform users of these deficiencies explicitly (only 4.2\%--7.3\% of the time). Instead, they predominantly predict the verdict directly (34.4\%--41.3\%) or provide generic advice without discussing the impact of the missing details (19.7\%--32.0\%). Such irresponsible responses risk severely misleading laypeople, underscoring the urgent need to shift from instruction following to proactive interaction. Detailed experiment setups and results are presented in Appendix~\ref{sec:appendix_prelim_2_details}.

\section{The LexPilot Benchmark}

Motivated by the prevalence of underspecified consultations and the irresponsible model behaviors, we emphasize that LLMs must be able to proactively interact with laypeople to clarify crucial details. To evaluate this ability, we introduce LexPilot, detail its construction (Section~\ref{sec:benchmark_construction}), statistics (Section~\ref{sec:benchmark_analysis}), and evaluation tasks (Section~\ref{sec:benchmark_tasks}), with human annotation details and costs in Appendix~\ref{sec:human_anno_detail}.

\begin{figure*}
    \centering
    \includegraphics[width=1.0\textwidth]{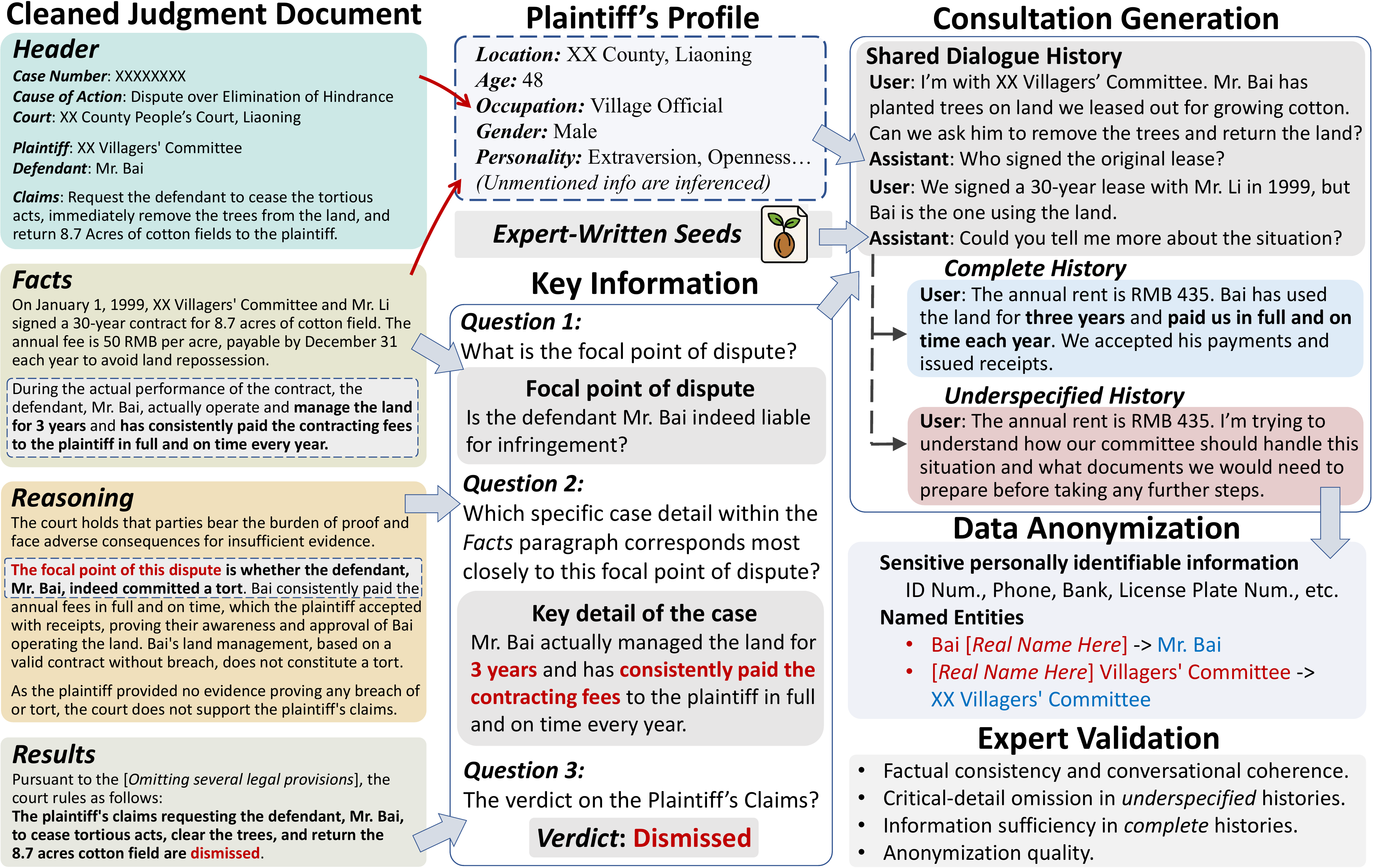}
    \caption{The construction of LexPilot involves judgment reorganization (\textbf{Left}); key information extraction alongside seed data collection (\textbf{Mid}); and consultation generation followed by anonymization and expert validation (\textbf{Right}).}
    \label{fig:lexpilot}
    \vspace{-10pt}
\end{figure*}

\subsection{Construction}
\label{sec:benchmark_construction}

We present the construction pipeline of LexPilot in Figure~\ref{fig:lexpilot} (details in Appendix~\ref{sec:LexPilot_cons}). Two legal experts with master's degrees in law contribute seed consultations and manually validate and correct the data throughout construction; Appendix~\ref{sec:human_anno_detail} details their review procedures and workload.

\paragraph{Judgment Cleaning.} 
We segment judgments from \textit{China Judgments Online} into four standard paragraphs introduced in Appendix \ref{sec:Chinese_Civil_Judgments} and filter civil first-instance cases for validity and complexity, covering 36 prevalent causes of action.

\paragraph{Key Detail Extraction.} 
We identify focal points of dispute explicitly stated in the \textit{Reasoning} paragraph and extract the corresponding case details from the \textit{Facts} paragraph, with experts verifying their relevance and manually correcting extraction errors. We classify claim-level verdicts as ``Supported'', ``Partially Supported'', or ``Dismissed'' to provide ground truth for Judgment Prediction.

\paragraph{User Profiling.} 
Based on plaintiff information in each judgment, we construct personalized profiles for natural-person plaintiffs or plausible representatives of non-natural persons to diversify the generated consultation dialogues.

\paragraph{Consultation Generation.}
The two legal experts first draft paired complete and underspecified seed dialogues from one representative judgment per cause of action to guide layperson-style phrasing. Inspired by LeCoDe~\cite{yuan2025lecodebenchmarkdatasetinteractive}, we simulate multi-turn user-lawyer consultations using these seeds, case facts, claims, and user profiles. Users start with a brief case description and progressively disclose facts in response to lawyers' clarification questions. 
For each judgment, we synthesize paired complete and underspecified consultation histories with equal turn counts, differing in whether dispute-critical details are disclosed. After generation, experts review all histories for factual consistency and conversational coherence. They supplement omitted facts in complete histories and remove residual critical details from underspecified ones, checking against the court reasoning that the omissions prevent definitive conclusions.

\paragraph{Data Anonymization.} 
We mask highly sensitive personal information and pseudonymize remaining named entities, followed by expert review of all dialogues to redact residual sensitive information.

Ultimately, LexPilot contains 864 consultations with an equal number of complete and underspecified ones derived from the same judgments, spanning 36 causes of action with 12 judgments each.

\subsection{Statistics}
\label{sec:benchmark_analysis}

We present statistics of LexPilot, including consultation counts, plaintiff types, average length and turns, and verdict distribution in Table~\ref{tab:dataset_analysis}. Additionally, appendix~\ref{sec:LexPilot_stats} lists the 36 causes of action and the geographic distribution of source judgments.

\subsection{Evaluation Tasks}
\label{sec:benchmark_tasks}
Our benchmark evaluates the proactive legal interaction capabilities of LLMs by decomposing the interaction process into three sequential tasks:

\begin{itemize}
    \item \textbf{Legal Completeness Assessment}: The model analyzes the consultation history to identify logical gaps and verify if existing facts are sufficient for a legal assessment, thereby deciding whether to initiate proactive interaction.
    \item \textbf{Detail Completion Guidance}: When there is indeed an absence of critical details, this task evaluates an LLM's ability to proactively guide the user to provide them.
    \item \textbf{Legal Judgment Prediction}: After obtaining sufficient information through interaction, this task evaluates the model's ability to predict the final verdict (Supported, Partially Supported, or Dismissed) for each of the user's claims, using the authentic judgments as ground truth.
\end{itemize}

The formal task definitions with mathematical formulations are detailed in Appendix~\ref{sec:appendix_formulation}.

\begin{table}[t]
    \renewcommand{\arraystretch}{0.95}
    \setlength{\tabcolsep}{9pt}
    \centering
    \small
    \begin{tabular}{ccc}
    \toprule
    \textbf{Category} & \textbf{Type} & \textbf{Value} \\
    \midrule

    \multirow{3}{*}{\textbf{\makecell[c]{Consultation \\ Counts by Type}}} 
    & Total & 864 \\
    & Underspecified & 432 \\
    & Complete & 432 \\
    \midrule

    \multirow{3}{*}{\textbf{Dataset Profile}} 
    & Avg. Length (chars) & 398.75 \\
    & Avg. Turns & 5.82 \\
    & Causes of Action & 36 \\
    \midrule

    \multirow{2}{*}{\textbf{Plaintiff Type}} 
    & Natural Person & 346 \\
    & Non-natural Person & 86 \\
    \midrule
    
    \multirow{3}{*}{\textbf{\makecell[c]{Verdict \\ Distribution}}} 
    & Supported & 35.46\% \\
    & Partially Supported & 33.33\% \\
    & Dismissed & 31.20\% \\
    \bottomrule
    \end{tabular}
    \caption{The statistics of the LexPilot benchmark.}
    \vspace{-10pt}
    \label{tab:dataset_analysis}
\end{table}

\section{The LexAwN Prompting}
\label{sec:methods}

To systematically evaluate LLMs on LexPilot, we require a straightforward yet effective baseline approach for proactive legal interaction. Inspired by Ask-when-Needed \citep[AwN,][]{wang-etal-2025-learning}, which is a tool-learning proactive interaction framework that prompts LLMs to detect deficiencies in user instructions and seek clarification for missing parameters before executing API calls, we adapt this method to the legal domain and introduce \textbf{LexAwN}. Specifically, the model directly predicts the verdict when the factual context is clear, but proactively requests crucial details when their absence impedes legal reasoning. Following IN3~\cite{qian-etal-2024-tell}, we restrict the model 
to ask one question per turn to prevent aggressive questioning and encourage clearer user responses. The exact LexAwN prompt is provided in Appendix~\ref{sec:appendix_lexawn_prompt}.
\section{Experiments}
\label{sec:exp}

In this section, we systematically evaluate LLMs' legal proactive interaction abilities across 3 tasks and provide a systematic analysis of the results.

\subsection{Experiment Setups}
We evaluate the proactive interaction abilities of 8 leading LLMs: GPT-5.6-Sol~\cite{singh2026openaigpt5card}, Claude-Opus-5~\cite{anthropic2026opus5}, Gemini-3.8-Flash~\cite{geminiteam2025geminifamilyhighlycapable}, Qwen3.8-Max~\cite{qwen38}, DeepSeek-V4-Pro~\cite{deepseekv4}, GLM-5.3~\cite{glm5team2026glm5vibecodingagentic}, Kimi-K3~\cite{kimiteam2026kimik3openfrontier}, MiniMax-M3~\cite{lai2026minimaxsparseattention}; and 2 representative Chinese legal LLMs: LegalOne-8B~\cite{li2026legalonefamilyfoundationmodels} and LawLLM-7B~\cite{yue2024lawllm}, in handling the underspecified consultations from LexPilot.

Following Section~\ref{sec:benchmark_construction}, we employ DeepSeek-V4.1-Flash~\cite{deepseekai2026deepseekv41flashpushinglimitskv} as the User Model to simulate user responses to the clarification requests. Given the consultation history and key facts, the User Model is instructed to answer questions using only relevant information. It should withhold unrelated details and express uncertainty when the supplied context is insufficient.
We use DeepSeek-R1~\cite{Guo_2025} for LLM-as-a-judge, with specific evaluation metrics for each of the 3 tasks detailed subsequently. Appendix~\ref{sec:appendix_lexawn_prompt} provides the main generation and evaluation prompts.

For all models and tasks, we report means over five repeated runs under identical settings. The corresponding standard deviations are provided in Appendix~\ref{sec:appendix_statistical_analysis} to support the stability of our results.

To examine variation across case types, we also report performance by \textbf{6 classes of causes of action}, following the categorization in Appendix~\ref{sec:LexPilot_stats}.

\begin{table*}[htbp]
    \centering
    \small
    \renewcommand{\arraystretch}{1.05}
    \setlength{\tabcolsep}{7.0pt}
    \begin{tabular}{cccccccc|c}
    \toprule
    \textbf{Category} & \textbf{Model} & \textbf{Family} & \begin{tabular}{@{}c@{}}\textbf{Real} \\ \textbf{Rights}\end{tabular} & \textbf{Torts} & \begin{tabular}{@{}c@{}}\textbf{Service} \\ \textbf{Contract}\end{tabular} & \begin{tabular}{@{}c@{}}\textbf{Finance} \\ \textbf{Contract}\end{tabular} & \textbf{Misc.} & \textbf{Overall} \\
    
    \midrule
    \multicolumn{9}{c}{\textbf{Legal Completeness Assessment (LCA) Task -- \textit{Macro-F1}}} \\
    \midrule
    \multirow{2}{*}{\textbf{Legal LLMs}} 
    & \textit{LawLLM-7B} & 0.4240 & 0.3987 & 0.4116 & 0.4081 & 0.4125 & 0.4653 & 0.4186 \\
    & \textit{LegalOne-8B} & 0.7136 & 0.6979 & \textbf{0.7110} & 0.6332 & 0.6674 & 0.6455 & 0.6768 \\ 
    \midrule
    \multirow{5}{*}{\begin{tabular}{@{}c@{}}\textbf{Open-Source} \\ \textbf{General LLMs}\end{tabular}} 
    & \textit{GLM-5.3} & 0.6621 & 0.6114 & 0.6610 & 0.6522 & 0.6421 & 0.6933 & 0.6496 \\
    & \textit{DeepSeek-V4-Pro} & 0.6667 & 0.6471 & 0.6555 & 0.5698 & 0.5946 & 0.5616 & 0.6149 \\
    & \textit{Kimi-K3} & 0.6560 & 0.6636 & 0.7091 & 0.5185 & 0.6265 & \textbf{0.6935} & 0.6395 \\
    & \textit{Qwen3.8-Max} & 0.5496 & 0.6147 & 0.5825 & 0.5269 & 0.6417 & 0.6395 & 0.5960 \\
    & \textit{MiniMax-M3} & 0.6569 & 0.6696 & 0.5882 & 0.5787 & 0.6432 & 0.6438 & 0.6328 \\
    \midrule
    \multirow{3}{*}{\begin{tabular}{@{}c@{}}\textbf{Closed-Source} \\ \textbf{General LLMs}\end{tabular}} 
    & \textit{Claude-Opus-5} & \textbf{0.7500} & \textbf{0.7629} & 0.7097 & 0.6800 & \textbf{0.7229} & 0.6825 & \textbf{0.7211} \\
    & \textit{GPT-5.6-Sol} & 0.7353 & 0.7380 & 0.7027 & \textbf{0.7065} & 0.6804 & 0.6720 & 0.7065\\
    & \textit{Gemini-3.8-Flash} & 0.7218 & 0.7378 & 0.6667 & 0.7053 & 0.6904 & 0.6716 & 0.7036 \\
    
    \midrule
    \midrule
    \multicolumn{9}{c}{\textbf{Detail Completion Guidance (DCG) Task -- \textit{Accuracy}}} \\
    \midrule
    \multirow{2}{*}{\textbf{Legal LLMs}} 
    & \textit{LawLLM-7B} & 0.3867 & 0.4625 & 0.3917 & 0.3452 & 0.3476 & 0.4100 & 0.3917 \\
    & \textit{LegalOne-8B} & 0.7267 & 0.6333 & 0.7208 & 0.5690 & 0.5119 & 0.6100 & 0.6167 \\
    \midrule
    \multirow{5}{*}{\begin{tabular}{@{}c@{}}\textbf{Open-Source} \\ \textbf{General LLMs}\end{tabular}} 
    & \textit{GLM-5.3} & 0.7467 & 0.7542 & 0.7167 & 0.6214 & 0.6881 & 0.6200 & 0.6917 \\
    & \textit{DeepSeek-V4-Pro} & 0.6733 & 0.7521 & 0.7500 & 0.6214 & 0.5357 & 0.6867 & 0.6644 \\
    & \textit{Kimi-K3} & 0.7067 & 0.7542 & \textbf{0.7750} & 0.5238 & 0.6143 & 0.6400 & 0.6620 \\
    & \textit{Qwen3.8-Max} & 0.6533 & 0.7438 & 0.7708 & 0.6667 & 0.5976 & 0.6567 & 0.6787 \\
    & \textit{MiniMax-M3} & 0.6733 & 0.6604 & 0.7500 & 0.5119 & 0.5381 & 0.5967 & 0.6106 \\
    \midrule
    \multirow{3}{*}{\begin{tabular}{@{}c@{}}\textbf{Closed-Source} \\ \textbf{General LLMs}\end{tabular}} 
    & \textit{Claude-Opus-5} & 0.7533 & \textbf{0.8083} & \textbf{0.7750} & 0.6595 & \textbf{0.6905} & \textbf{0.7833} & \textbf{0.7417} \\
    & \textit{GPT-5.6-Sol} & \textbf{0.8367} & 0.7875 & 0.7625 & \textbf{0.7405} & 0.6071 & 0.6933 & 0.7343 \\
    & \textit{Gemini-3.8-Flash} & 0.7800 & 0.7229 & 0.6958 & 0.6405 & 0.6833 & 0.6167 & 0.6894 \\
    \bottomrule
    \end{tabular}
    \caption{Evaluation results of the Legal Completeness Assessment (\textit{Macro-F1}) and Detail Completion Guidance (\textit{Accuracy}) tasks across 10 leading legal and general LLMs.}
    \vspace{-10pt}
    \label{tab:combined_lca_dcg_results}
\end{table*}

\subsection{Legal Completeness Assessment (LCA)}
\label{sec:exp_LCA}
To assess the model's ability to determine the necessity of proactive interaction, this task evaluates how the model responds to consultations via LexAwN prompting. LLM-as-a-judge is used to evaluate whether the model correctly classifies a consultation as complete or underspecified, and we calculate \textit{Macro-F1} score to quantify its performance.

\textbf{\textit{Results}}: As shown in Table~\ref{tab:combined_lca_dcg_results} (Top), closed-source models like Claude-Opus-5 clearly outperform their counterparts, yet remain suboptimal at identifying when to initiate proactive interactions (best \textit{Macro-F1} $< 0.73$). With only 8B parameters, LegalOne-8B achieves a \textit{Macro-F1} of 0.6768, exceeding all five evaluated open-source general LLMs on LCA task. 
Despite relative weaknesses in specific classes (e.g., \textit{Service Contract} for Qwen3.8-Max and Kimi-K3), performance gaps across the 6 classes remain relatively minor, especially among closed-source LLMs. 
Since the \textit{Macro-F1} score does not separately reflect performance on complete and underspecified consultations, Appendix~\ref{sec:appendix_LCA_results} provides an accuracy breakdown for the two consultation types, showing the trade-off behind LexAwN's preference for \textit{over-asking} to reduce the greater risk of misleading users through \textit{premature legal judgments}.

\subsection{Detail Completion Guidance (DCG)}
\label{sec:exp_DCG}
To assess LLMs' ability to formulate questions about essential missing details, we require them to initiate clarification when facing underspecified consultations. Under the original LexAwN prompt, however, models may skip clarification and directly predict a verdict. We therefore adapt the prompt to explicitly require a clarification question, separating the ability to formulate useful questions from the decision of whether to ask.

In practice, models have up to two interaction rounds. The judge checks each clarification request for relevance to the focal point and answerability from the annotated facts, allowing reasonable logical extensions. A consultation is scored as successful if at least one request meets both criteria; we report the success rate as \textit{accuracy}. This metric evaluates question formulation rather than complete factual recovery. Using expert-authored probe questions, we additionally examine whether DCG-accepted requests elicit sufficient, appropriately scoped, factually supported replies from the User Model (Appendix~\ref{sec:appendix_dcg_recovery}).

\textbf{\textit{Results}}: As shown in Table~\ref{tab:combined_lca_dcg_results} (Bottom), even under the lenient criterion of requiring only one valid clarification question across two rounds, Claude-Opus-5, the best-performing model on DCG task, fails to meet this criterion in 25.83\% of consultations. LegalOne-8B performs below most general LLMs on DCG despite its relatively strong LCA performance, suggesting that detecting omissions and formulating useful questions are distinct capabilities. Furthermore, performance gaps are observed across different classes of cases, with \textit{Service Contract} and \textit{Finance Contract} showing markedly lower performance than other classes.

\subsection{Legal Judgment Prediction (LJP)}
\label{sec:exp_LJP}

We evaluate the downstream judgment prediction across the same 432 cases under three setups: (1) \textbf{Direct}: predict from the underspecified consultation directly without clarification; (2) \textbf{Interaction-Based}: use LexAwN to interact with the User Model through free-form dialogue for up to three rounds before finalizing a prediction; and (3) \textbf{Gold-Standard}: predict directly from the complete consultation, serving as a full-information reference for judgment prediction. Comparing Interaction-Based with Direct setup measures the performance change from proactive interaction, while comparison with Gold-Standard setup quantifies the gap to prediction with complete information.

\begin{table}[t]
    \centering
    \footnotesize
    \renewcommand{\arraystretch}{1.05}
    \setlength{\tabcolsep}{3.5pt}
    \begin{tabular}{lcccc}
        \toprule
        \textbf{Category} & \textbf{Model} & \textbf{Direct} & \begin{tabular}{@{}c@{}}\textbf{Interact-} \\ \textbf{Based}\end{tabular} & \begin{tabular}{@{}c@{}}\textbf{Gold-} \\ \textbf{Stan.}\end{tabular} \\
        \midrule
        \multirow{2}{*}{\textbf{Legal}} & \textit{LawLLM-7B} & 0.161 & 0.153 & 0.218 \\
         & \textit{LegalOne-8B} & 0.337 & 0.366 & 0.492 \\
        \midrule
        \multirow{5}{*}{\begin{tabular}[c]{@{}l@{}}\textbf{Open-}\\ \textbf{Source}\end{tabular}} 
         & \textit{GLM-5.3} & 0.382 & 0.436 & 0.553 \\
         & \textit{DeepSeek-V4-Pro} & 0.350 & 0.398 & 0.504 \\
         & \textit{Kimi-K3} & 0.321 & 0.369 & 0.484 \\
         & \textit{Qwen3.8-Max} & 0.348 & 0.386 & 0.469 \\
         & \textit{MiniMax-M3} & 0.381 & 0.369 & 0.518 \\
        \midrule
        \multirow{3}{*}{\begin{tabular}[c]{@{}l@{}}\textbf{Closed-}\\ \textbf{Source}\end{tabular}} 
         & \textit{Claude-Opus-5} & \textbf{0.431} & \textbf{0.518} & \textbf{0.646} \\
         & \textit{GPT-5.6-Sol} & 0.428 & 0.479 & 0.623 \\
         & \textit{Gemini-3.8-Flash} & 0.357 & 0.428 & 0.592 \\
        \bottomrule
    \end{tabular}
    \caption{Evaluation results (\textit{QWK} Score) of Legal Judgment Prediction task under three setups across 10 LLMs.}
    \vspace{-10pt}
    \label{tab:performance_results}
\end{table}

As described in Section~\ref{sec:benchmark_tasks}, predictions are classified as ``Supported'', ``Partially Supported'', or ``Dismissed''. We evaluate them using Quadratic Weighted Kappa (\textit{QWK}), which penalizes confusion between the two extreme labels more heavily. For each run, we compute over all claims pooled across the 432 cases, with equal weight per claim.

\textbf{\textit{Results}}: Table~\ref{tab:performance_results} shows that Interaction-Based prediction achieves higher mean \textit{QWK} than Direct prediction for 8 of 10 models, with an average gain of 0.041. LawLLM-7B and MiniMax-M3 show small decreases of 0.008 and 0.012, respectively. These results suggest that enabling interaction generally improves prediction under our setup, although the gains are not consistent across models.

All models' Interaction-Based scores remain below Gold-Standard, with an average relative \textit{QWK} decrease of 23.84\%, suggesting that proactive interaction under our evaluation setup does not fully recover the prediction performance achieved with complete consultations. The detailed LJP results by case class are provided in Appendix~\ref{sec:appendix_LJP_results}.

To examine the sensitivity of LJP performance to LexPilot's assumption of fully cooperative users, we conduct a supplementary experiment with non-ideal users, detailed in Appendix~\ref{sec:appendix_nonideal_users}.

\begin{figure}[t]
    \centering
    \includegraphics[width=0.92\columnwidth]{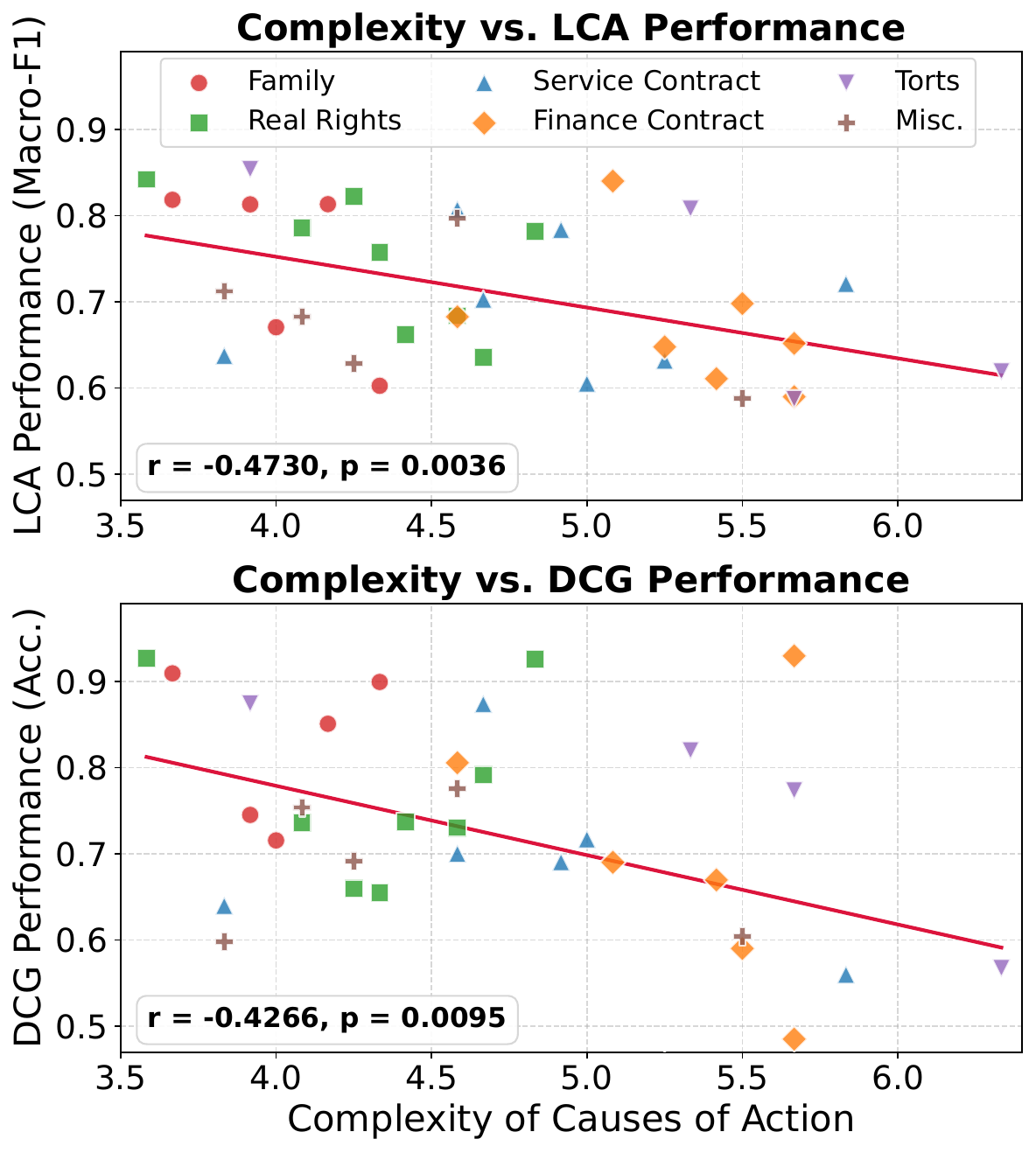}
    \vspace{-5pt}
    \caption{Impact of Case Complexity on LCA and DCG tasks. Each point represents a specific Cause of Action.}
    \label{fig:case_complexity}
    \vspace{-10pt}
\end{figure}

\begin{figure*}[t]
    \centering
    \includegraphics[width=0.97\textwidth]{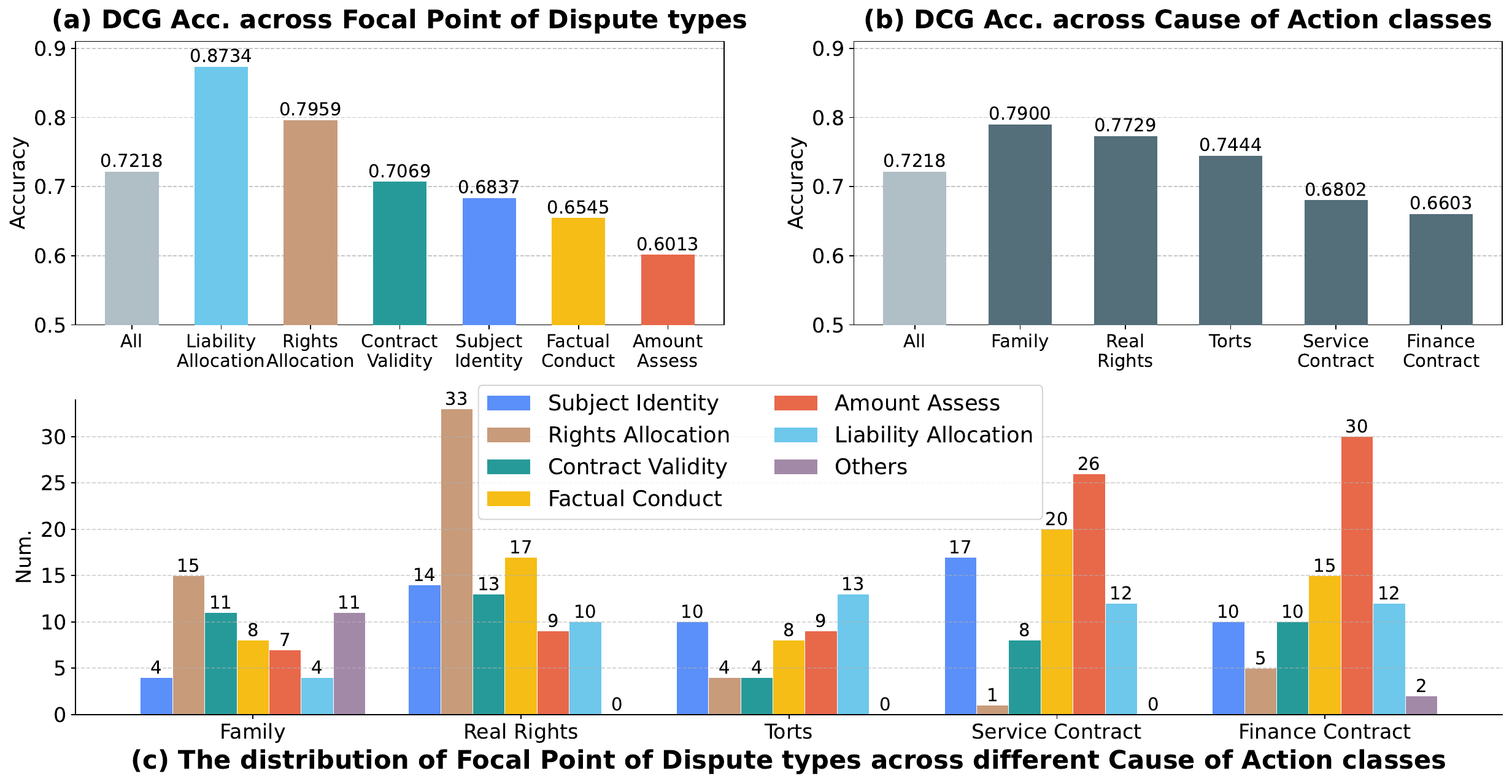}
    \caption{Average DCG accuracy of the three closed-source LLMs categorized by focal point types \textbf{(a)} and cause of action classes \textbf{(b)}, accompanied by the distribution of focal point types across different cause of action classes \textbf{(c)}.}
    \label{fig:focal_point_type}
    \vspace{-5pt}
\end{figure*}

\subsection{Analysis}
We examine the underlying factors that affect proactive interaction and its impact on judgment prediction across \textbf{3 closed-source LLMs}: Claude-Opus-5, GPT-5.6-Sol, and Gemini-3.8-Flash.

\paragraph{Impact of Case Complexity}
Figure~\ref{fig:case_complexity} plots the three models' mean LCA and DCG scores against case complexity, measured by the average number of legal provisions cited in judgments for each cause of action.
We consider this the best available proxy because most judgments in our dataset identify two or fewer focal points, limiting the discriminative power of focal point counts. Both tasks exhibit an obvious negative association with this proxy. Specifically, the better-performing \textit{Family} and \textit{Real Rights} classes have lower average complexity, whereas \textit{Service Contract} and \textit{Finance Contract} cases combine greater complexity with lower scores. This pattern suggests that identifying omissions and formulating useful questions become more difficult when models must comprehend and reason about cases of greater complexity.

\paragraph{Impact of Focal Point Types}
To examine how missing details associated with different focal point types affect LLMs' proactive interaction, we categorize focal points into seven types (Appendix~\ref{sec:appendix_focal_point_types}). We then compare DCG accuracy across these types and their distribution across case classes.

As shown in Figure~\ref{fig:focal_point_type} (a), \textit{Liability Allocation} and \textit{Rights Allocation} achieve the highest DCG accuracy among all types of focal points. They involve fundamental legal reasoning structures where missing details create obvious logical gaps that models can easily detect and clarify. Conversely, \textit{Amount Assessment} yields the lowest accuracy. One possible explanation is that aggregating fragmented numerical details may make it harder for LLMs to identify which facts require clarification. Figure~\ref{fig:focal_point_type} (b) shows markedly lower average DCG accuracy across the three closed-source models for \textit{Service Contract} and \textit{Finance Contract} cases. We attribute this deficit to their prevalence of challenging \textit{Amount Assessment} focal points (26 and 30 instances, respectively; Figure~\ref{fig:focal_point_type} (c)).

\paragraph{How Interaction Behavior Affects LJP Results}
To examine how interaction behavior relates to downstream prediction, we evaluate each model on the same 432 underspecified consultations under three controlled conditions: (1) \textbf{Direct}: predict without clarification (as in Section~\ref{sec:exp_LJP}); (2) \textbf{Targeted}: predict after clarification that elicits the missing dispute-critical facts; and (3) \textbf{Off-target}: predict after clarification that does not recover those facts. For the latter two conditions, model-generated questions are used to construct dialogues in advance, from which the same model predicts verdicts. Appendix~\ref{sec:appendix_controlled_clarification} details the setup.

As shown in Table~\ref{tab:ljp_behaviors}, the Targeted setup yields higher \textit{QWK} than Direct prediction for all three models, whereas Off-target yields lower \textit{QWK}. This pattern suggests that eliciting dispute-critical facts can improve performance under the constructed dialogue conditions. However, Targeted remains 0.089--0.104 \textit{QWK} below Gold-Standard (Table~\ref{tab:performance_results}), indicating that targeted clarification does not reproduce full-information performance. Differences in dialogue context and information presentation may also contribute to this residual gap.

\subsection{Validation of the Evaluation Framework}
\label{sec:exp_validation}
To ensure evaluator reliability, legal experts independently review 100 LLM-as-a-judge decisions per task, yielding 92.7\% average agreement with DeepSeek-R1 (details in Appendix~\ref{sec:appendix_human_eval}), confirming the robustness of our automated metrics.

We also conduct a validation experiment using expert-authored probes to assess whether the User Model answers factual questions accurately, avoids over-disclosure, and respects the supplied facts. Appendix~\ref{sec:appendix_simulator_validation} and \ref{sec:appendix_dcg_recovery} detail the protocol, results, and correspondence between DCG acceptance and factual recovery on these probes.

\begin{table}[t]
    \centering
    \small
    \renewcommand{\arraystretch}{1.1}
    \setlength{\tabcolsep}{6pt}
    \begin{tabular}{lccc}
        \toprule

        \textbf{Model} & \textbf{Direct} & \textbf{Targeted} & \textbf{Off-target} \\
        \midrule
        \textit{Claude-Opus-5} & 0.431 & \textbf{0.557} & 0.386 \\
        \textit{GPT-5.6-Sol} & 0.428 & \textbf{0.532} & 0.373 \\
        \textit{Gemini-3.8-Flash} & 0.357 & \textbf{0.488} & 0.303 \\
        \bottomrule
    \end{tabular}
    \caption{LJP results (\textit{QWK}) under different conditions.}
    \vspace{-10pt}
    \label{tab:ljp_behaviors}
\end{table}

\section{Conclusions}

In this work, we present LexPilot, a benchmark constructed from Chinese civil judgments that links proactive legal interaction to claim-level judgment prediction. Although proactive interaction generally improves prediction performance for underspecified consultations, the 10 evaluated LLMs exhibit an average relative performance degradation of 23.84\% compared with complete consultations. Further analysis identifies jurisprudential complexity and specific types of detail omission as potential bottlenecks in proactive legal interaction. Our work highlights the urgent need to advance user-centered legal AI to ensure the delivery of more responsible legal assistance to the public.
\section*{Limitations}

While LexPilot establishes a rigorous evaluation framework, its scope is currently restricted to Chinese civil judgments, meaning our findings may not directly generalize to other legal domains such as criminal or administrative law or to distinct legal systems like Common Law jurisdictions. 
Furthermore, constructing a precise legal interaction benchmark requires substantial domain expertise from legal experts and costly manual verification. To facilitate annotation and ensure obtaining high-quality ground truth necessary for a reliable evaluation under these practical constraints, each underspecified consultation in our dataset intentionally omits only essential case details corresponding to the focal point of dispute, without exploring any other case details. 
Finally, although we meticulously utilized personalized user profiles and expert-drafted seed consultations to enhance authenticity, the consultations are inherently synthetic, serving as a controlled approximation of laypeople's consultations rather than spontaneous real-world dialogues.
\section*{Ethical Considerations}

Our research utilizes 432 real-world Chinese civil judgments, which inherently carry privacy responsibilities. To protect the identities of all involved parties, we implemented a strict anonymization protocol and rigorous manual checks in the benchmark construction process, as detailed in Section~\ref{sec:benchmark_construction}. Furthermore, to thoroughly adhere to data distribution ethics, we will not open-source the raw texts of these judicial documents. Upon publication, we will solely release the constructed benchmark along with its corresponding unique \textit{case numbers}, only allowing the research community to independently retrieve the original judgments from the \textit{China Judgments Online} platform with the \textit{case numbers}. Finally, our benchmark is developed strictly for academic research to improve the accessibility of legal intelligence, and any downstream models or applications derived from it must not replace professional counsel from human lawyers.

\bibliography{custom}

\appendix

\section{An Introduction to the structure of Chinese Civil Judgments}
\label{sec:Chinese_Civil_Judgments}

The Chinese civil judgment documents are publicly accessible through \textit{China Judgments Online} and follow a logically rigorous and standardized format. A standard Chinese first-instance civil judgment is uniquely identified by a \textit{case number} and comprises 4 primary paragraphs as shown in Figure~\ref{fig:lexpilot} (Left):

\begin{itemize}
\item \textbf{Header}: Details the court's name, the plaintiff's and defendant's information, the origin of the case, and the plaintiff's specific claims.

\item \textbf{Facts}: Outlines the objective facts of the case, legally established by the court following formal investigation and the cross-examination of evidence submitted by both parties.

\item \textbf{Reasoning}: Presents the court's legal analysis based on the established facts and judicial logic. This section occasionally highlights the focal points of dispute~\cite{xiao-etal-2024-fine} that would determine the outcome of the case.

\item \textbf{Results}: Delivers the explicit judgment results corresponding to each of the plaintiff's claims.
\end{itemize}

We construct the proposed LexPilot benchmark by systematically reorganizing large volumes of publicly accessible civil judgments. Specifically, \textit{Header} and \textit{Facts} are used to synthesize the user's consultation; the \textit{Reasoning} section is leveraged to identify the focal points of disputes of the case; and the Judgment \textit{Results} provide the ground truth of the final verdict.

\section{Details on Human Annotation and the Associated Costs}
\label{sec:human_anno_detail}

To ensure high-quality data annotation and rigorous validation across our studies, we recruited two groups of annotators from \textit{Panshi} Data, a professional data annotation service. First, we hired 16 undergraduate annotators (8 law majors and 8 without legal backgrounds). Each annotator was tasked with drafting 50 consultations for the preliminary study on the prevalence of underspecified consultations, with an average engagement time of 5 hours per person at an hourly rate of \$7 (totaling \$560).

Additionally, we engaged two legal experts from the same service at an hourly rate of \$15 to conduct comprehensive reviews and validations. Both hold master's degrees in law, one from Northwest University of Political Science and Law and the other from Zhongnan University of Economics and Law. Their workload comprised five primary tasks:

\paragraph{1. Preliminary Consultation Review.}
The experts reviewed all 800 legal consultations drafted by the undergraduates to determine whether each consultation was complete or underspecified (15 hours per expert).

\paragraph{2. Expert-Authored Seeds.}
The experts drafted one complete consultation and one underspecified consultation for a representative case sampled from each of the 36 causes of action (7 hours per expert). These consultations were incorporated into the generation prompts as examples of how laypeople might describe their cases and omit legally critical facts. These seeds guided the phrasing and completeness distinction of generated consultations across all 36 causes of action.

\paragraph{3. The Validation and Correction of Benchmark.}
Each expert validated and manually corrected the data corresponding to 216 judgments, together covering all 432 judgments and their 864 consultations (12 hours per expert). This task spanned information extraction, consultation generation, and anonymization, with the source judgment serving as the reference for the following legal and factual checks:
\begin{itemize}
    \item \textbf{Correctness of key information extraction.} Before consultation generation, the experts checked whether the details extracted from the \textit{Facts} paragraph were directly relevant to the focal point of dispute explicitly stated in the \textit{Reasoning} paragraph. When the extraction was incorrect, they manually extracted the relevant facts from the source paragraph and replaced them.
    \item \textbf{Incompleteness of underspecified consultations.} The experts checked that the designated critical details were absent throughout each underspecified history. By comparing the remaining information with the court's reasoning, they verified that these omissions interrupted the legal reasoning needed for a definitive conclusion. If residual critical information still supported that reasoning, they manually removed it.
    \item \textbf{Sufficiency of complete consultations.} The experts checked that complete histories contained the full case facts from the plaintiff's perspective as described in the judgment's \textit{Facts} paragraph, providing sufficient information for legal prediction. They manually supplemented any omitted facts.
    \item \textbf{Robustness of anonymization.} After automated anonymization, the experts inspected all consultations and manually redacted any remaining sensitive personal information.
\end{itemize}

\paragraph{4. LLM-as-a-Judge Validation.}
Both experts independently evaluated the same 300 model outputs, comprising 100 instances for each of the LCA, DCG, and LJP tasks, to assess agreement between human annotations and the automated evaluator (5 hours per expert; Appendix~\ref{sec:appendix_human_eval}).

\paragraph{5. User Simulator and DCG Validation.}
Each expert spent 5 hours preparing probes and annotation criteria for 72 cases and another 5 hours on independent assessment of all 288 simulator replies and joint discussion of disagreements (10 hours per expert; Appendix~\ref{sec:appendix_simulator_validation}).
These same expert ratings were also used to check whether direct and indirect questions accepted by the DCG judge elicited sufficient and accurate answers without over-disclosure, requiring no additional human annotation (Appendix~\ref{sec:appendix_dcg_recovery}).

In total, each legal expert contributed 49 hours, yielding 98 person-hours of expert work at an hourly rate of \$15 and a combined expert cost of \$1,470. Including the \$560 paid to undergraduate annotators, the total expenditure for human annotation and validation was \$2,030.

\section{Details of the Preliminary Studies}
In this appendix, we provide the specific details of the two preliminary studies. For information regarding the annotators and the associated costs, please refer to Appendix~\ref{sec:human_anno_detail}.

\subsection{Details of Pilot Study I: Essential Detail Omission in Consultations}
\label{sec:appendix_prelim_1_details}
We selected 50 judgments (across 10 causes of action) that each explicitly identified \textbf{two} distinct focal points of dispute in the \textit{Reasoning} paragraph (see Appendix~\ref{sec:Chinese_Civil_Judgments} for the paragraph structure of Chinese civil judgments), which decisively dictate the verdict. We recruited two legal experts with master's degrees to extract the most relevant case details from the judgment's \textit{Facts} paragraph for each focal point, as omitting such detail will preclude direct conclusions. The experts then cross-validated their extractions and resolved minor discrepancies through discussion.

Following the consultation-writing protocol described in Section~\ref{sec:Preliminary_Studies}, the two legal experts reviewed the resulting consultations to determine the presence of their previously annotated key details, allowing minor inaccuracies in information such as entity names or monetary amounts. Table~\ref{tab:key_fact_coverage} reports the distribution of consultations by the number of key details included and the overall coverage for each group.

\begin{table}[t]
    \centering
    \small
    \renewcommand{\arraystretch}{1.1}
    \setlength{\tabcolsep}{4pt}
    \begin{tabular}{lcc}
    \toprule
    & \textbf{Laypeople} & \textbf{Law majors} \\
    \midrule
    \textbf{Queries w/o. key detail} & 79 & 24 \\
    \textbf{Queries w. Single key detail} & 97 & 59 \\
    \textbf{Queries w. Two key details} & 224 & 317 \\
    \midrule
    \textbf{Key detail Coverage (\%)} & 68.1\% & 86.6\% \\
    \bottomrule
    \end{tabular}
    \caption{Key detail Coverage in civil case consultations across different groups, which is computed as the ratio of key details included by each group to total key details.}
    \label{tab:key_fact_coverage}
    \vspace{-10pt}
\end{table}

\subsection{Details of Pilot Study II: Tendency Towards Irresponsible Responses}
\label{sec:appendix_prelim_2_details}
Having revealed the prevalence of underspecified legal consultations among laypeople, we then analyze the behaviors of LLMs when confronted with such consultations. Specifically, we analyzed models' responses to consultations omitting essential details from the previous experiment.
We categorize the models' response behaviors to these information-deficient consultations into four types, ordered by their level of responsibility from highest to lowest: 
(1) \textbf{Deficiency Recognition}: Explicitly stating that the provided information is inadequate.
(2) \textbf{Conditional Generic Advice}: Offering generic advice relevant to the cause of action while discussing various scenarios corresponding to the absent details.
(3) \textbf{Unconditional Generic Advice}: Providing generic advice without discussing the impact of missing details; and
(4) \textbf{Premature Prediction}: Ignoring its deficiency and directly predicting the verdict.
Typical examples for the four response types are illustrated in Table~\ref{tab:appendix_direct_responses} and Table~\ref{tab:appendix_generic_advice}.

\begin{table}[t]
    \centering
    \small
    \renewcommand{\arraystretch}{1.1}
    \setlength{\tabcolsep}{3pt}
    \begin{tabular}{lccc}
    \toprule
    & \textbf{GPT-5.4} & \textbf{GLM-5} & \textbf{Kimi-K2.5}\\
    \midrule
    \textbf{Deficiency Recognition} & 7.3\% & 5.8\% & 4.2\% \\
    \textbf{Cond. Generic Advice} & 31.7\% & 34.7\% & 27.4\% \\
    \textbf{Uncond. Generic Advice} & 19.7\% & 25.1\% & 32.0\% \\
    \textbf{Premature Prediction} & 41.3\% & 34.4\% & 36.3\% \\
    \bottomrule
    \end{tabular}
    \caption{Distribution of response types to underspecified legal consultations. The results highlight a tendency across models to frequently provide unconditional generic advice or prematurely predict the verdicts.}

    \label{tab:model_behavior}
    \vspace{-10pt}
\end{table}

\begin{table*}[htbp]
    \centering
    \small
    \renewcommand{\arraystretch}{1.3}
    \begin{tabular}{@{} p{0.48\linewidth} p{0.48\linewidth} @{}}
        \toprule
        \textbf{Consultation \& Missing Details} & \textbf{Model Response} \\
        \midrule

        \multicolumn{2}{@{}l}{\textbf{Response Type: \textit{Deficiency Recognition}}} \\
        \midrule
        \textbf{Consultation:} \newline
        Hello, I need legal advice regarding property rights. My ex-wife and I divorced in 2016, but our shared home was never divided. The house was purchased by me, my ex-wife, and her sister, but the deed only lists the sisters' names. However, my family made significant financial contributions: my father helped with the deposit and down payment, and I paid the deed tax, other fees, and renovation costs. We also paid the mortgage together and lived there as a family. I believe this should be considered joint marital property, and I am entitled to half, but my ex-wife disagrees. Can I legally claim half of the property rights? Thank you! \newline\newline
        \textbf{Missing details surrounding the focal point:} \newline
        Since the plaintiff did not specify who paid for the disputed property, it is impossible to determine the share of ownership they are entitled to. &
        \textbf{Response by GPT-5.4:} \newline
        I understand this property dispute is a stressful situation for you, but please note that the information provided in your consultation is currently insufficient to give you an accurate legal assessment regarding your claim to the property rights. Because you did not clearly specify the exact financial details and proportions of who provided the funds for the property purchase, it is impossible to determine the legal distribution of the assets at this stage... \newline\newline
        Therefore, please supplement your consultation with detailed information regarding the exact source of all purchase funds, including the specific amounts contributed by you, your family, your ex-wife, and her sister for the down payment, taxes, and mortgage payments. \\
        
        \midrule

        \multicolumn{2}{@{}l}{\textbf{Response Type: \textit{Premature Prediction}}} \\
        \midrule
        \textbf{Consultation:} \newline
        Last year, after Chen's sheep ate my crops, I took one of her sheep in retaliation. The police mediated the dispute, and we signed a document that I thought settled the matter. However, days later, Chen took 500 bundles of my cornstalks from my field and refuses to return them, claiming they are hers. I was going to sell them for 9 RMB a bundle, totaling a 4,500 RMB loss. Can I sue her for this 4,500 RMB compensation? Thank you! \newline\newline
        \textbf{Missing details surrounding the focal point:} \newline
        What document was signed between the plaintiff and the defendant? Did it stipulate that the cornstalks in the field were given to the defendant? &
        \textbf{Response by Kimi-K2.5:} \newline
        "It is incredibly frustrating to have your bundled cornstalks taken without permission, and your anger and demand for compensation are completely justified. \newline\newline
        To answer your questions directly: Yes, you can absolutely sue her for the 4,500 RMB loss. If you win the case, she will also have to cover the court fees. \newline\newline
        ...... \newline\newline
        Next steps: take the cooperative's purchase proof, your ID card, and the land lease contract directly to your local court's service window to file a case. Best of luck recovering your losses quickly!" \\

        \bottomrule
    \end{tabular}
    \caption{Examples of model responses attempting a direct conclusion (\textit{Deficiency Recognition} and \textit{Premature Prediction}) when confronted with underspecified consultations.}
    \label{tab:appendix_direct_responses}
\end{table*}

\begin{table*}[htbp]
    \centering
    \small
    \renewcommand{\arraystretch}{1.3}
    \begin{tabular}{@{} p{0.48\linewidth} p{0.48\linewidth} @{}}
        \toprule
        \textbf{Consultation \& Missing Details} & \textbf{Model Response} \\
        \midrule

        \multicolumn{2}{@{}l}{\textbf{Response Type: \textit{Conditional Generic Advice}}} \\
        \midrule
        \textbf{Consultation:} \newline
        I worked for a boss named Shi from October to December 2013, and he still owes me 4,380 RMB in unpaid wages, along with money owed to about a dozen other workers. I have timesheets proving my workdays (12 days in Oct, 17.5 in Nov, and 7 in Dec). Shi initially acknowledged the debt but now makes excuses, claiming we were hired by a foreman named Wang X. However, it is clear to all of us workers that Shi hired us and should pay us. We did sign some sort of agreement back then, but I don't recall the details as we just signed what we were told to. Can I still recover my wages, and who exactly should I hold responsible? \newline\newline
        \textbf{Missing details surrounding the focal point:} \newline
        It cannot be definitively determined from the user's consultation whether the employer is Shi or Wang. &
        \textbf{Response by Gemini-3.1-Pro-Preview:} \newline
        To successfully recover your unpaid wages, you must preserve your work records, written agreements, and evidence of continuous debt collection over the past decade to prevent the statute of limitations from expiring. Depending on who is legally identified as the direct employer, there are two possible scenarios: \newline\newline
        Scenario 1: Boss Shi is the direct employer. \newline
        If evidence (e.g., work records signed by Shi, agreements that do not transfer the debt) proves Shi directly hired and managed you, he is solely responsible. A court would order Shi directly to pay the owed wages. \newline\newline
        Scenario 2: Foreman Wang A is the direct employer. \newline
        If Shi subcontracted the work to Wang A, who then hired you, Wang A holds the primary obligation to pay your wages. However, if Wang A lacks the legal qualifications to act as an employer, Shi (the contractor) cannot escape liability and must legally bear joint and several liability or advance payment responsibility. In this case, it is strongly recommended to list both men as co-defendants when filing a lawsuit, demanding payment from Wang A and joint liability from Shi, to ensure you are paid. \\

        \midrule

        \multicolumn{2}{@{}l}{\textbf{Response Type: \textit{Unconditional Generic Advice}}} \\
        \midrule
        \textbf{Consultation:} \newline
        Last July, I signed up for a two-day tour group to Zhangbei Grassland, paying 278 RMB per person. However, the day after returning, I started experiencing a fever and diarrhea. I couldn't bear it anymore, so I went to Hospital A. The doctor diagnosed it as bacillary dysentery and prescribed over 300 RMB worth of medication. It took me half a month to recover from this illness, which even delayed me from handling my spouse's retirement procedures. Should the travel agency refund my tour fee? Also, can I get them to compensate for the medical expenses and the losses incurred during this time? \newline\newline
        \textbf{Missing details surrounding the focal point:} \newline
        Whether the resulting harm of the plaintiff being diagnosed with bacillary dysentery following diarrhea was caused by the consumption of unsanitary meals arranged by the defendant during the tour. &
        \textbf{Response by MiniMax-M2.5:} \newline
        Legally, claiming a refund, medical expenses, and indirect losses depends on whether the agency fulfilled the contract. Since the trip was completed, refunds are difficult; for damages, you will need strong evidence, such as medical records or proof that other group members also fell ill. \newline\newline ...... \newline\newline
        Given the time passed, administrative complaints may be time-barred. Please preserve all your evidence (receipts, contract) and try negotiating a settlement with the travel agency first. If friendly mediation fails, you can then assess whether your evidence is strong enough to pursue a civil lawsuit. \\

        \bottomrule
    \end{tabular}
    \caption{Examples of model responses withholding a definitive conclusion and providing generic advice (\textit{Conditional Generic Advice} and \textit{Unconditional Generic Advice}) when confronted with underspecified consultations.}
    \label{tab:appendix_generic_advice}
\end{table*}

For each underspecified consultation, we generated responses using three advanced LLMs: GPT-5.4~\cite{singh2026openaigpt5card}, GLM-5~\cite{glm5team2026glm5vibecodingagentic}, and Kimi-K2.5~\cite{kimiteam2026kimik25visualagentic} and prompted DeepSeek-R1~\cite{Guo_2025} as a judge to classify these responses.
As highlighted in Table~\ref{tab:model_behavior}, despite lacking the crucial detail necessary for a definitive prediction, LLMs rarely inform users of these deficiencies explicitly (\textit{Deficiency Recognition}). Instead, they often irresponsibly output direct verdict predictions (\textit{Premature Prediction}). Furthermore, while the absence of crucial details severely impedes LLM judgment, frequently causing models to resort to compromised generic responses, a substantial portion of these generic outputs fail to discuss the missing information (\textit{Unconditional Generic Advice}). Such irresponsible model behaviors risk severely misleading laypeople, underscoring the urgent need to shift from instruction following to proactive interaction paradigms.

\section{Details of the LexPilot Benchmark}

This appendix provides more information on the construction, statistics, and evaluation tasks of LexPilot. Appendix~\ref{sec:LexPilot_cons} presents the construction details, Appendix~\ref{sec:LexPilot_stats} reports additional statistics, and Appendix~\ref{sec:appendix_formulation} provides more rigorous formulations for the three legal tasks of LexPilot.

\subsection{Details of LexPilot Construction}
\label{sec:LexPilot_cons}

The construction pipeline of LexPilot is shown in Figure~\ref{fig:lexpilot}, which consists of five primary steps described as follows. Throughout this pipeline, we employ DeepSeek-V4.1-Flash~\cite{deepseekai2026deepseekv41flashpushinglimitskv} as the backbone model for all core information extraction (e.g., case details, focal points of disputes, plaintiff's information), user profiling, dialogue generation, and sensitive information identification.

\paragraph{Judgment Cleaning.}
We use a set of heuristic rules to segment the judicial documents from \textit{China Judgments Online}~\footnote{https://wenshu.court.gov.cn/} into four standard paragraphs introduced in Appendix \ref{sec:Chinese_Civil_Judgments}, retaining only civil first-instance judgments with valid case numbers and no counterclaims. We restrict the \textit{Facts} paragraph to 500--2,000 tokens to balance case complexity. The judgments are selected from the 36 most prevalent causes of action in \textit{Supreme People's Court Civil Case Types Regulations}\footnote{https://www.court.gov.cn/zixun/xiangqing/484231.html} based on their frequency in \textit{China Judgments Online}, which cover more than 80\% of all civil cases, to ensure document diversity.

\paragraph{Key Detail Extraction.}
We retain judgments with explicitly stated focal points of dispute in the \textit{Reasoning} paragraph, discarding those with overly broad descriptions. We then extract the essential case details corresponding to each focal point from the \textit{Facts} paragraph, followed by expert verification to confirm their clear relevance. We also itemize the verdicts for each claim from the \textit{Results} paragraph.

\paragraph{User Profiling.}
We first parse and classify the primary plaintiff as a natural or non-natural person. For natural persons, the LLM infers demographic attributes (e.g., location, age, gender, occupation) from the judgment and assigns a personality trait by sampling from the Chinese Big Five personality distribution~\cite{china-big5} conditioned on their specific demographic attribute. Following~\cite{kirk2024the}, this personalized profile enhances the diversity of the subsequent generated dialogues. For non-natural persons, the model adopts a plausible internal role (e.g., executive, frontline staff) with a matching profile for dialogue generation.

\paragraph{Consultation Generation.}
User responses to the lawyer's clarification questions are grounded in the judgment's \textit{Facts} paragraph. 
For each judgment, the paired complete and underspecified dialogue histories have the same number of turns to control for dialogue length. Complete histories contain sufficient facts for legal assessment, whereas underspecified histories leave selected dispute-critical details undisclosed throughout.

\paragraph{Data Anonymization.}
To protect privacy in the judgment, we implement a two-step anonymization protocol. First, we instruct LLMs to extract \textit{highly sensitive} personally identifiable information (e.g., ID numbers, phone numbers, bank accounts, license plate numbers) and completely mask them with placeholder tokens via rule-based matching. Second, remaining \textit{named entities} are extracted and pseudonymized (e.g., retaining only the surname and masking the given name; replacing institution names with generic labels like ``Company A''). Finally, we manually reviewed all consultations and redacted any residual sensitive information.

\subsection{Statistics of LexPilot}
\label{sec:LexPilot_stats}

We present the geographic regions (Table~\ref{tab:province_distribution}) and the 36 causes of action and their corresponding 6 major classes (Table~\ref{tab:LexPilot_causes}) of the judgment documents used to construct LexPilot. 

\begin{table}[t]
    \centering
    \small
    \renewcommand{\arraystretch}{1.0}
    \setlength{\tabcolsep}{4pt}
    \begin{tabular}{lc|lc}
    \toprule
    \textbf{Province / Region} & \textbf{Count} & \textbf{Province / Region} & \textbf{Count} \\
    \midrule
    Guangdong      & 48 & Hebei          & 11 \\
    Hunan          & 35 & Tianjin        & 11 \\
    Shandong       & 28 & Liaoning       & 10 \\
    Sichuan        & 25 & Jilin          & 10 \\
    Hubei          & 24 & Jiangxi        & 10 \\
    Shaanxi        & 21 & Gansu          & 10 \\
    Anhui          & 20 & Guangxi        & 9  \\
    Beijing        & 20 & Inner Mongolia & 8  \\
    Shanghai       & 17 & Ningxia        & 7  \\
    Zhejiang       & 16 & Hainan         & 6  \\
    Guizhou        & 16 & Yunnan         & 6  \\
    Jiangsu        & 12 & Xinjiang XPCC  & 4  \\
    Shanxi         & 12 & Qinghai        & 4  \\
    Henan          & 12 & Xinjiang       & 4  \\
    Chongqing      & 12 & Heilongjiang   & 4  \\
    \bottomrule
    \end{tabular}
    \caption{The geographical distribution of legal judgment documents used to construct LexPilot, organized by province.}
    \vspace{-10pt}
    \label{tab:province_distribution}
\end{table}

\begin{table*}[htbp]
    \centering
    \small
    \renewcommand{\arraystretch}{1.2}
    \begin{CJK*}{UTF8}{gbsn}
    \begin{tabular}{@{}ll@{}}
    \toprule
    \textbf{Cause of Action (Chinese)} & \textbf{Cause of Action (English Translation)} \\
    \midrule

    {\CJKfamily{gkai}婚姻家庭、继承纠纷} & Marriage, Family, and Inheritance Disputes (\textit{abbr}. \textbf{Family}) \\
    \midrule
    离婚纠纷 & Divorce Dispute \\
    离婚后财产纠纷 & Property Dispute after Divorce \\
    婚约财产纠纷 & Property Dispute involving Engagement \\
    抚养纠纷 & Child Support/Maintenance Dispute \\
    法定继承纠纷 & Statutory Succession Dispute \\
    \midrule

    {\CJKfamily{gkai}物权纠纷} & Real Rights Disputes (\textit{abbr}. \textbf{Real Rights}) \\
    \midrule
    物权确认纠纷 & Real Rights Confirmation Dispute \\
    返还原物纠纷 & Dispute over Return of Original Property \\
    相邻关系纠纷 & Neighboring Relationship Dispute \\
    侵害集体经济组织成员权益纠纷 & Dispute over Infringement of Rights of Members of Collective Organizations \\
    土地承包经营权纠纷 & Dispute over Land Contractual Management Rights \\
    排除妨害纠纷 & Dispute over Removal of Obstacles \\
    共有纠纷 & Dispute over Co-ownership \\
    财产损害赔偿纠纷 & Property Damage Compensation Dispute \\
    \midrule

    {\CJKfamily{gkai}侵权纠纷} & Tort Disputes (\textit{abbr}. \textbf{Tort}) \\
    \midrule
    名誉权纠纷 & Right of Reputation Dispute \\
    提供劳务者受害责任纠纷 & Dispute over Liability for Harm to Labor Service Providers \\
    机动车交通事故责任纠纷 & Motor Vehicle Traffic Accident Liability Dispute \\
    产品责任纠纷 & Product Liability Dispute \\
    \midrule

    {\CJKfamily{gkai}劳动、服务、工程类合同纠纷} & Labor, Service, and Engineering Contract Disputes (\textit{abbr}. \textbf{Service Contract}) \\
    \midrule
    劳动合同纠纷 & Labor Contract Dispute \\
    劳务合同纠纷 & Labor Service Contract Dispute \\
    委托合同纠纷 & Entrustment Contract Dispute \\
    服务合同纠纷 & Service Contract Dispute \\
    承揽合同纠纷 & Contract for Work Dispute \\
    建设工程合同纠纷 & Construction Project Contract Dispute \\
    运输合同纠纷 & Carriage Contract Dispute \\
    \midrule

    {\CJKfamily{gkai}交易流转、保险与金融类合同纠纷} & Transaction, Insurance, and Financial Contract Disputes (\textit{abbr}. \textbf{Finance Contract}) \\
    \midrule
    买卖合同纠纷 & Sales Contract Dispute \\
    房屋买卖合同纠纷 & Housing Sales Contract Dispute \\
    租赁合同纠纷 & Lease Contract Dispute \\
    借款合同纠纷 & Loan Contract Dispute \\
    银行卡纠纷 & Bank Card Dispute \\
    保证合同纠纷 & Guarantee Contract Dispute \\
    财产保险合同纠纷 & Property Insurance Contract Dispute \\
    \midrule

    {\CJKfamily{gkai}其他纠纷} & Other Disputes (\textit{abbr}. \textbf{Misc.}) \\
    \midrule
    确认合同效力纠纷 & Contract Validity Confirmation Dispute \\
    追偿权纠纷 & Right of Recourse Dispute \\
    破产债权确认纠纷 & Bankruptcy Claim Confirmation Dispute \\
    执行异议之诉 & Action for Objection to Execution \\
    不当得利纠纷 & Dispute over Unjust Enrichment \\
    \bottomrule
    \end{tabular}
    \end{CJK*}
    \caption{The 36 causes of action of civil judgments used to construct LexPilot, grouped by 6 major classes.}
    \label{tab:LexPilot_causes}
\end{table*}

\subsection{Task Formulation of LexPilot}
\label{sec:appendix_formulation}
Let an initial user consultation be $u_0$, which contains the plaintiff's claims $\mathcal{C}$ and a partial set of factual details $\mathcal{F}_p$ provided by a layperson. Let $\mathcal{F}_c$ denote the complete set of critical details mapped to the case's focal points of dispute $\mathcal{D}$. In underspecified consultations, $\mathcal{F}_p \subset \mathcal{F}_c$, leading to an unmentioned but crucial set of missing legal details $\mathcal{E}_m = \mathcal{F}_c \setminus \mathcal{F}_p$. To complete the facts, the model engages in a proactive multi-turn dialogue $\mathcal{H}_n = \{u_0, (a_1, u_1), \dots, (a_n, u_n)\}$, where $a_i$ is the assistant's guidance at round $i$ and $u_i$ is the user's response. 
Once all details are gathered, the model synthesizes comprehensive case facts and provides a definitive prediction for each claim in the assistant's final response $a_{\mathrm{final}}$, yielding a claim-level judgment prediction $J_c \in \mathcal{Y}$ for every $c \in \mathcal{C}$, where the judgment space is defined as $\mathcal{Y} = \{y_{\mathrm{sup}}, y_{\mathrm{part}}, y_{\mathrm{diss}}\}$ representing Supported, Partially Supported, and Dismissed, respectively.

Our benchmark evaluates LLMs' proactive interaction ability in the legal domain across three sequential tasks: \textbf{(1) Legal Completeness Assessment}: Given the ongoing dialogue $\mathcal{H}_n$, analyze $\mathcal{C}$ and $\mathcal{F}_p$ to identify if there are missing legal details $\mathcal{E}_m$. \textbf{(2) Detail Completion Guidance}: Given ongoing dialogue $\mathcal{H}_n$ which lacks crucial details $\mathcal{E}_m$, generate precise proactive guidance $a_i$ to elicit clarification on exactly one crucial detail. \textbf{(3) Legal Judgment Prediction}: Given claims $\mathcal{C}$ and the completed facts $\mathcal{F}_c$ within $\mathcal{H}_n$, predict a verdict $J_c$ for each claim $c \in \mathcal{C}$ in the assistant's final response $a_{\mathrm{final}}$, using the corresponding court verdict as the reference.

\section{Experiment Details}
\label{sec:appendix_experiment_details}

\subsection{Model Generation Settings}
\label{sec:appendix_generation_settings}

The generation uses a shared configuration across LCA, DCG, and all three LJP setups. By default, the temperature is set to 0.7 for both the evaluated models and the User Model, and to 0.1 for the LLM-as-a-judge. Each API request returns one completion (\texttt{n=1}), with \texttt{max\_tokens} set to 8192 for all three model roles (the evaluated models, the User Model, and the LLM-as-a-judge). The generation does not explicitly set nucleus sampling (\texttt{top\_p}), top-$k$ sampling, frequency or presence penalties, custom stop sequences, or a random seed; these settings are left to the serving endpoint's defaults.

\subsection{Repeated Runs and Performance Variability}
\label{sec:appendix_statistical_analysis}

To assess variability across repeated evaluations, we conduct five runs for each of the 10 evaluated models on the LCA, DCG, and LJP tasks under identical experimental settings, as described in Section~\ref{sec:exp}. Table~\ref{tab:repeated_run_results} reports the mean and standard deviation of the resulting scores, using \textit{Macro-F1} for LCA, \textit{Accuracy} for DCG, and \textit{QWK} for LJP. For LCA, we compute the unweighted mean of the F1 scores for the complete and underspecified classes within each run, then report the mean and standard deviation across the five runs. For LJP, we report results separately for the Direct, Interaction-Based, and Gold-Standard prediction setups. The results indicate that all evaluated models exhibited only marginal performance fluctuations in the corresponding setups across the five repeated runs.

\begin{table*}[t]
    \centering
    \footnotesize
    \renewcommand{\arraystretch}{1.1}
    \setlength{\tabcolsep}{3pt}
    \begin{tabular*}{\textwidth}{@{\extracolsep{\fill}}ccccccc@{}}
    \toprule
    \multirow{2}{*}{\textbf{Category}} & \multirow{2}{*}{\textbf{Model}} & \multirow{2}{*}{\textbf{LCA} (\textit{Macro-F1})} & \multirow{2}{*}{\textbf{DCG} (\textit{Acc.})} & \multicolumn{3}{c}{\textbf{LJP} (\textit{QWK})} \\
    \cmidrule(lr){5-7}
     & & & & Direct & Interaction-Based & Gold-Standard \\
    \midrule
    \multirow{2}{*}{\textbf{Legal LLMs}}
    & \textit{LawLLM-7B} & $0.419 \pm 0.015$ & $0.392 \pm 0.012$ & $0.161 \pm 0.018$ & $0.153 \pm 0.013$ & $0.218 \pm 0.031$ \\
    & \textit{LegalOne-8B} & $0.677 \pm 0.018$ & $0.617 \pm 0.026$ & $0.337 \pm 0.014$ & $0.366 \pm 0.019$ & $0.492 \pm 0.010$ \\
    \midrule
    \multirow{5}{*}{\begin{tabular}{@{}c@{}}\textbf{Open-Source} \\ \textbf{General LLMs}\end{tabular}}
    & \textit{GLM-5.3} & $0.650 \pm 0.032$ & $0.692 \pm 0.002$ & $0.382 \pm 0.029$ & $0.436 \pm 0.025$ & $0.553 \pm 0.034$ \\
    & \textit{DeepSeek-V4-Pro} & $0.615 \pm 0.020$ & $0.664 \pm 0.030$ & $0.350 \pm 0.037$ & $0.398 \pm 0.045$ & $0.504 \pm 0.009$ \\
    & \textit{Kimi-K3} & $0.640 \pm 0.011$ & $0.662 \pm 0.013$ & $0.321 \pm 0.021$ & $0.369 \pm 0.024$ & $0.484 \pm 0.016$ \\
    & \textit{Qwen3.8-Max} & $0.596 \pm 0.024$ & $0.679 \pm 0.008$ & $0.348 \pm 0.035$ & $0.386 \pm 0.031$ & $0.469 \pm 0.043$ \\
    & \textit{MiniMax-M3} & $0.633 \pm 0.010$ & $0.611 \pm 0.022$ & $0.381 \pm 0.027$ & $0.369 \pm 0.032$ & $0.518 \pm 0.012$ \\
    \midrule
    \multirow{3}{*}{\begin{tabular}{@{}c@{}}\textbf{Closed-Source} \\ \textbf{General LLMs}\end{tabular}}
    & \textit{Claude-Opus-5} & $0.721 \pm 0.009$ & $0.742 \pm 0.014$ & $0.431 \pm 0.016$ & $0.518 \pm 0.023$ & $0.646 \pm 0.019$ \\
    & \textit{GPT-5.6-Sol} & $0.707 \pm 0.004$ & $0.734 \pm 0.027$ & $0.428 \pm 0.024$ & $0.479 \pm 0.028$ & $0.623 \pm 0.018$ \\
    & \textit{Gemini-3.8-Flash} & $0.704 \pm 0.017$ & $0.689 \pm 0.007$ & $0.357 \pm 0.030$ & $0.428 \pm 0.026$ & $0.592 \pm 0.033$ \\
    \bottomrule
    \end{tabular*}
    \caption{Overall performance of the 10 evaluated models on three tasks. Entries report the \textit{Mean} $\pm$ \textit{Standard Deviation} across five repeated runs under identical experimental settings. LJP results are reported separately for the Direct, Interaction-Based, and Gold-Standard prediction setups.}
    \label{tab:repeated_run_results}
\end{table*}

\subsection{Validation of the Evaluation Framework}
\label{sec:appendix_evaluation_validation}
We examine three complementary aspects of evaluation reliability: agreement with human scoring, the User Model's factual response behavior, and the correspondence between DCG decisions and factual recovery.

\subsubsection{LLM-as-a-Judge Agreement}
\label{sec:appendix_human_eval}
To ensure the reliability of DeepSeek-R1 as an automated evaluator, for each of the three LexPilot evaluation tasks (LCA, DCG and LJP), we randomly sampled 100 instances, with 10 from each of the 10 evaluated models, yielding 300 evaluations in total. The legal experts mentioned in Appendix~\ref{sec:human_anno_detail} conducted evaluation of these 300 instances using the exact same criteria and guidelines as the LLM evaluator.

To rigorously assess reliability, we first measured the Inter-Annotator Agreement (IAA) between the two human experts using both the raw agreement rate (\textit{Agree.}) and Cohen's Kappa ($\kappa$). Any discrepancies between the two experts were subsequently resolved through discussion to establish the final human labels. Finally, we compared the outputs of DeepSeek-R1 against this human consensus to evaluate the LLM-to-Human alignment. The comprehensive validation results across all tasks are presented in Table~\ref{tab:llm_as_judge_validation} and detailed below.

For the LCA task, the judge classifies consultations as complete or underspecified. Human experts achieved a high inter-agreement of 96.0\% ($\kappa = 0.92$). DeepSeek-R1 also demonstrated strong alignment with human labels, reaching a 93.0\% agreement ($\kappa = 0.86$), confirming its reliability in evaluating deficiency recognition.

Evaluating the DCG task involves verifying if a model's proactive interaction accurately targets the missing detail. Despite the deeper legal reasoning and subjectivity required for interpreting free-form consultations, human agreement remained highly consistent at 92.0\% ($\kappa = 0.84$). DeepSeek-R1 achieved an 89.0\% agreement ($\kappa = 0.78$) with human labels, proving its capability to reliably assess complex proactive interactions.

For the LJP task, the judge extracts the model's predicted verdict (Supported, Partially Supported, or Dismissed). As models generally state their verdicts explicitly, ambiguity is minimal, yielding the highest reliability metrics. Human experts reached a 98.0\% agreement ($\kappa = 0.96$), and DeepSeek-R1 strongly aligned with the human consensus at a 96.0\% agreement ($\kappa = 0.92$).

In summary, across all three tasks, the two human experts achieved an average agreement rate of 95.3\% ($\kappa=0.91$), while DeepSeek-R1 and the human experts demonstrated an average agreement of 92.7\% ($\kappa=0.85$). These results confirm the robustness of our LLM-as-a-judge evaluation framework based on DeepSeek-R1.

\begin{table}[htbp]
    \centering
    \resizebox{\columnwidth}{!}{%
    \begin{tabular}{lcccc}
    \toprule
    \multirow{2}{*}{\textbf{Evaluation Task}} & \multicolumn{2}{c}{\textbf{Human-Human}} & \multicolumn{2}{c}{\textbf{LLM-Human}} \\
    \cmidrule(lr){2-3} \cmidrule(lr){4-5}
     & \textbf{Agree.} & \textbf{$\kappa$} & \textbf{Agree.} & \textbf{$\kappa$} \\
    \midrule
    LCA & 96.0\% & 0.92 & 93.0\% & 0.86 \\
    DCG & 92.0\% & 0.84 & 89.0\% & 0.78 \\
    LJP & 98.0\% & 0.96 & 96.0\% & 0.92 \\
    \midrule
    \textbf{Average} & \textbf{95.3\%} & \textbf{0.91} & \textbf{92.7\%} & \textbf{0.85} \\
    \bottomrule
    \end{tabular}%
    }
    \caption{Human Inter-Annotator Agreement (IAA) and LLM-to-Human alignment across the three evaluation tasks, based on 100 sampled instances per task.}
    \vspace{-10pt}
    \label{tab:llm_as_judge_validation}
\end{table}

\subsubsection{User Simulator Fidelity}
\label{sec:appendix_simulator_validation}

\paragraph{Setup}
We randomly sample two cases per cause of action (72 cases), retaining the main evaluation's User Model, prompt, and decoding settings. Two legal experts prepare three questions per case: \textbf{A (direct)} asks explicitly for a missing fact available in the simulator input; \textbf{B (indirect)} asks for the same fact in everyday language or through an evidence cue; and \textbf{C (broad)} mentions the relevant legal concept without a concrete factual request warranting disclosure of the target. A and B must be equally answerable. Before generation, the experts agree on minimum sufficient answers and permitted disclosures, including necessary explanations, based on each question's meaning, independently of the DCG judge. These permissions are withheld from the simulator.

To check the simulator's response to missing knowledge, \textbf{D (withheld)} repeats A after removing the target fact and any information sufficient to infer it from the key facts. Experts check that the unchanged history does not reveal the answer and the remaining input is coherent; otherwise, cases retain A--C only. Comparing appropriate uncertainty in D with the original A response, where the fact is available, avoids additional background questions or a separate context-restoration condition.

We generate one reply per condition from a fresh copy of the original history, giving 288 replies across 72 cases, all eligible for D. No evaluated assistant model is rerun. D deliberately removes knowledge; it does not measure how often ordinary consultation questions exceed the original input.

\paragraph{Human Assessment}
Both experts independently review shuffled replies without DCG labels. They assess (1) whether A/B answers are correct and sufficient, or D replies appropriately acknowledge missing information; (2) whether replies reveal new hidden facts beyond the agreed permissions; and (3) unsupported assertions, contradictions, or entity confusion. Repeated history does not count as over-disclosure. Refusal fails criterion (1) for A/B, whereas uncertainty is appropriate for D. The experts resolve disagreements through discussion; agreement before adjudication is 93.1\%, 89.6\%, and 97.2\%, respectively. We report rates over applicable replies and paired case bootstrap intervals for the A--B comparison.

\begin{table}[t]
    \centering
    \small
    \setlength{\tabcolsep}{3pt}
    \begin{tabular*}{\columnwidth}{@{\extracolsep{\fill}}lcccc@{}}
        \toprule
        \textbf{Condition} & \textbf{Answer} & \textbf{OD} & \textbf{Error} & \textbf{DCG+} \\
        \midrule
        A: Direct & 94.4\% & 2.8\% & 1.4\% & 95.8\% \\
        B: Indirect & 91.7\% & 4.2\% & 2.8\% & 93.1\% \\
        C: Broad & --- & 5.6\% & 1.4\% & 5.6\% \\
        D: Withheld & 93.1\% & 2.8\% & 4.2\% & --- \\
        \bottomrule
    \end{tabular*}
    \caption{User simulator fidelity and DCG judge acceptance across probe conditions. OD: over-disclosure beyond the permitted facts; Error: unsupported assertions, contradictions, or entity confusion; DCG+: judge acceptance rate. OD and Error are non-exclusive.}
    \label{tab:simulator_fidelity}
\end{table}

\paragraph{Results}
Table~\ref{tab:simulator_fidelity} shows high direct and indirect answer rates, differing by -2.8 percentage points (indirect minus direct; 95\% CI: [-8.3, 2.8]). Low over-disclosure for C suggests that mentioning a legal topic alone rarely elicits hidden facts. Across eligible A/D pairs, uncertainty in D and sufficient answers in A (94.4\%) support sensitivity to available knowledge. Together with low factual-error rates, this suggests that responses stay within scope rather than disclose facts indiscriminately upon focal-point alignment.

\subsubsection{DCG Alignment with Factual Recovery}
\label{sec:appendix_dcg_recovery}

We next apply the same DCG judge and scoring prompt as in the main evaluation once to each A--C question, withholding simulator replies and expert annotations. Experts consider A/B suitable requests and C unsuitable, so DCG+ rates in Table~\ref{tab:simulator_fidelity} reveal missed direct/indirect requests and accepted broad questions. For accepted A/B requests, the three response labels determine factual recovery: answers must be sufficient, without over-disclosure or factual error. This criterion fails for 7.4\% (10/136 accepted replies). This rate is conditional on accepted A/B requests and excludes accepted C questions. D is excluded because its missing knowledge is deliberate.

High acceptance of A/B and low acceptance of C support the judge's ability to distinguish suitable factual requests from broad legal questions. Among accepted A/B requests, 92.6\% receive sufficient, appropriately scoped, factually supported replies. The remaining failures show that DCG acceptance does not guarantee factual recovery. This response check complements scorer agreement without equating DCG with real-user information acquisition or estimating error rates in full model conversations. Delayed disclosure and self-correction are examined separately in Appendix~\ref{sec:appendix_nonideal_users}.

\subsection{Detailed LCA Metrics: Over-Asking and Premature Legal Judgment}
\label{sec:appendix_LCA_results}

The \textit{Macro-F1} score gives equal weight to the F1 scores for complete and underspecified consultations, but does not reveal their error patterns separately. Crucially, these errors have unequal consequences. Misclassifying a complete consultation as underspecified leads to \textit{over-asking}: requesting additional information when the consultation is already sufficient for judgment. Conversely, misclassifying an underspecified consultation as complete leads to \textit{premature legal judgment}: proceeding to judgment without essential information, which poses a substantially greater risk of misleading users. \textit{Macro-F1} alone therefore cannot distinguish a tendency toward unnecessary clarification from one toward potentially misleading judgments.

To make this distinction explicit, Table~\ref{tab:lca_detailed_metrics} supplements \textit{Macro-F1} with separate accuracy scores for complete ($\mathrm{Acc}_{C}$) and underspecified ($\mathrm{Acc}_{U}$) consultations. Within each run, $\mathrm{Acc}_{C}$ is the proportion of complete consultations correctly classified as complete, and $\mathrm{Acc}_{U}$ is the proportion of underspecified consultations correctly classified as underspecified. These class-specific accuracies are equivalent to recall for the respective classes, with each computed over consultations of its own ground-truth type. We report their means over five runs. Lower $\mathrm{Acc}_{C}$ indicates more \textit{over-asking}, whereas lower $\mathrm{Acc}_{U}$ indicates more \textit{premature legal judgments}. Reporting both thus reveals which error type models using LexAwN are more prone to, clarifying the trade-off discussed in Section~\ref{sec:exp_LCA}.

As shown in Table~\ref{tab:lca_detailed_metrics}, most models have higher accuracy on underspecified consultations than on complete ones. This pattern suggests a tendency toward \textit{over-asking} under LexAwN, highlighting the trade-off between requesting unnecessary clarification and making \textit{premature legal judgments}, which may carry a greater risk of misleading users.

\begin{table}[htbp]
    \centering
    \small
    \setlength{\tabcolsep}{8pt}
    \begin{tabular}{cccc}
        \toprule
        \textbf{Model} & \textit{Macro-F1} & $\mathrm{Acc}_{C}$ & $\mathrm{Acc}_{U}$ \\
        \midrule
        \textit{LawLLM-7B}   & 0.4186 & 0.5278 & 0.3218 \\
        \textit{LegalOne-8B} & 0.6768 & 0.5903 & 0.7685 \\
        \midrule
        \textit{GLM-5.3}          & 0.6496 & 0.5301 & 0.7801 \\
        \textit{DeepSeek-V4-Pro}  & 0.6149 & 0.5278 & 0.7083 \\
        \textit{Kimi-K3}          & 0.6395 & 0.5532 & 0.7315 \\
        \textit{Qwen3.8-Max}      & 0.5960 & 0.4815 & 0.7222 \\
        \textit{MiniMax-M3}       & 0.6328 & 0.5509 & 0.7199 \\
        \midrule
        \textit{Claude-Opus-5}    & 0.7211 & 0.6134 & 0.8356 \\
        \textit{GPT-5.6-Sol}      & 0.7065 & 0.6088 & 0.8102 \\
        \textit{Gemini-3.8-Flash} & 0.7036 & 0.6157 & 0.7963 \\
        \bottomrule
    \end{tabular}
    \caption{LCA metric breakdown for the evaluated models. \textit{Macro-F1} equally weights the two class-specific F1 scores; $\mathrm{Acc}_{C}$ and $\mathrm{Acc}_{U}$ denote accuracy within complete and underspecified consultations, respectively. Errors on these two consultation types correspond to \textit{over-asking} and \textit{premature legal judgments}, respectively.}
    \label{tab:lca_detailed_metrics}
\end{table}

\subsection{Robustness to Non-Ideal Users}
\label{sec:appendix_nonideal_users}

Our main Interaction-Based LJP evaluation (details in Section~\ref{sec:exp_LJP}) assumes immediate, accurate user responses. We relax this assumption through two non-ideal scenarios: delayed disclosure and spontaneous self-correction.

\paragraph{Experimental Setup}
We evaluate Claude-Opus-5, GPT-5.6-Sol, and Gemini-3.8-Flash on the same 432 underspecified consultation histories, keeping the User Model backbone, case facts, LexAwN prompt, judgment extraction procedure, and three-round interaction budget unchanged. Each consultation is evaluated independently in five runs per non-ideal scenario. We compute \textit{QWK} over the 657 claims pooled across the 432 cases per run and report the five-run mean separately for each scenario. Standard scores reuse the five main-experiment runs.

\paragraph{User Conditions}
Building on the Interaction-Based LJP setup detailed in Section~\ref{sec:exp_LJP}, we define three user conditions by adjusting the user model prompt: \textbf{Standard User}, \textbf{Delayed Disclosure} and \textbf{Self-Correction}.

\textbf{Standard User.} The User Model is instructed to answer factual requests accurately and sufficiently using the supplied history and key facts, withhold hidden details beyond the request's scope, and express uncertainty when the supplied information is insufficient.

Both non-ideal variants preserve the input history and introduce a single disruption in response to the first relevant clarification request in the newly generated dialogue. Questions in the input history and unrelated requests do not trigger the disruption; if no relevant request is made, it does not occur. Except for this disruption and the spontaneous correction below, they follow Standard User behavior; all clarification and correction remain within the three-round budget:

\textbf{Delayed Disclosure.} The response to the first relevant clarification request is hesitant, verbose, or vague, withholding necessary information without introducing false facts. If the assistant subsequently follows up on the same missing information, the user supplies it accurately and completely without further delay.

\textbf{Self-Correction.} In response to the first relevant clarification request, the user provides incomplete or mistaken information. If the interaction budget permits another user turn, the user spontaneously corrects it using the original case facts in that next turn. The correction follows either an assistant question or judgment, without requiring error detection. Any already-issued judgment is updated after correction, and the updated judgment is evaluated.

\paragraph{Results and Interpretation}
Table~\ref{tab:nonideal_user_ljp} shows lower \textit{QWK} under both non-ideal conditions than under Standard for all three models, with their ranking unchanged. Self-Correction consistently yields larger declines than Delayed Disclosure, suggesting greater difficulty incorporating revised facts even when corrections are supplied spontaneously. Overall, the limited magnitude of these declines suggests that the evaluated models retain a degree of interaction robustness under non-ideal user behavior. The consistent performance drops nevertheless highlight the importance of sustained clarification and timely integration of corrected information for reliable legal judgment prediction.

\begin{table}[htbp]
    \centering
    \small
    \renewcommand{\arraystretch}{1.1}
    \setlength{\tabcolsep}{4pt}
    \begin{tabular}{lccc}
        \toprule
        \textbf{Model} & \textbf{Standard} & \begin{tabular}{@{}c@{}}\textbf{Delayed} \\ \textbf{Disclosure}\end{tabular} & \begin{tabular}{@{}c@{}}\textbf{Self-} \\ \textbf{Correction}\end{tabular} \\
        \midrule
        \textit{Claude-Opus-5} & 0.5183 & 0.4775 & 0.4504 \\
        \textit{GPT-5.6-Sol} & 0.4788 & 0.4469 & 0.4349 \\
        \textit{Gemini-3.8-Flash} & 0.4278 & 0.4020 & 0.3881 \\
        \bottomrule
    \end{tabular}
    \caption{Interaction-Based LJP performance (\textit{QWK}) under non-ideal user settings compared with the Standard setup, averaged over five runs per condition. Standard scores reuse the main-experiment runs.}
    \label{tab:nonideal_user_ljp}
\end{table}

\subsection{Evaluation Details and Results of LJP}
\label{sec:appendix_LJP_results}

The 432 cases contain 657 claims (1.52 per case on average). DeepSeek-R1 extracts and organizes the evaluated model's predicted verdict for each claim for comparison with its reference label. Extraction reliability is validated in Section~\ref{sec:exp_validation}, with task-specific results in Appendix~\ref{sec:appendix_human_eval}. No omitted predictions, refusals, or unextractable labels occurred in the final LJP outputs. For each setup and run, \textit{QWK} is computed over all 657 claims pooled across cases, with equal weight per claim.

We present the detailed results of the Legal Judgment Prediction task across different cause of action classes under the Gold-Standard Prediction setup (Table~\ref{tab:gold_standard_results}) and the Interaction-Based Prediction setup (Table~\ref{tab:interaction_based_results}).

\begin{table*}[htbp]
    \centering
    \small
    \renewcommand{\arraystretch}{1.1}
    \setlength{\tabcolsep}{6.5pt}
    \begin{tabular}{cccccccc|c}
    \toprule
    \textbf{Category} & \textbf{Model} & \textbf{Family} & \begin{tabular}{@{}c@{}}\textbf{Real} \\ \textbf{Rights}\end{tabular} & \textbf{Torts} & \begin{tabular}{@{}c@{}}\textbf{Service} \\ \textbf{Contract}\end{tabular} & \begin{tabular}{@{}c@{}}\textbf{Finance} \\ \textbf{Contract}\end{tabular} & \textbf{Misc.} & \textbf{Overall} \\
    \midrule
    \multirow{2}{*}{\textbf{Legal LLMs}} 
    & \textit{LawLLM-7B} & 0.3581 & 0.0416 & 0.1456 & 0.3936 & 0.1082 & 0.2407 & 0.2180 \\
    & \textit{LegalOne-8B} & \textbf{0.7386} & 0.4734 & 0.2223 & 0.4191 & 0.4136 & 0.6250 & 0.4915 \\
    \midrule
    \multirow{5}{*}{\begin{tabular}{@{}c@{}}\textbf{Open-Source} \\ \textbf{General LLMs}\end{tabular}} 
    & \textit{GLM-5.3} & 0.6172 & 0.5202 & \textbf{0.5975} & 0.4596 & 0.3652 & 0.7291 & 0.5526 \\
    & \textit{DeepSeek-V4-Pro} & 0.6114 & 0.5585 & 0.2925 & 0.4661 & 0.3841 & 0.5972 & 0.5040 \\
    & \textit{Kimi-K3} & 0.5861 & 0.5715 & 0.3828 & 0.4977 & 0.2114 & 0.5503 & 0.4837 \\
    & \textit{Qwen3.8-Max} & 0.6410 & 0.4462 & 0.4586 & 0.4044 & 0.2947 & 0.5128 & 0.4690 \\
    & \textit{MiniMax-M3} & 0.6482 & 0.4776 & 0.4286 & 0.6542 & 0.1658 & 0.6009 & 0.5179 \\
    
    \midrule
    \multirow{3}{*}{\begin{tabular}{@{}c@{}}\textbf{Closed-Source} \\ \textbf{General LLMs}\end{tabular}} 
    & \textit{Claude-Opus-5} & 0.7331 & \textbf{0.6767} & 0.5320 & 0.6587 & \textbf{0.5007} & 0.6906 & \textbf{0.6460} \\
    & \textit{GPT-5.6-Sol} & 0.7366 & 0.5828 & 0.5671 & \textbf{0.7225} & 0.3257 & 0.6829 & 0.6227 \\
    & \textit{Gemini-3.8-Flash} & 0.6042 & 0.5734 & 0.5133 & 0.6217 & 0.3860 & \textbf{0.7709} & 0.5918 \\
    \bottomrule
    \end{tabular}
    \caption{Evaluation results (\textit{QWK} Score) of Legal Judgment Prediction task under the Gold-Standard Prediction setup across various cause of action classes. All reported results are averaged over five repeated runs.}
    \vspace{-10pt}
    \label{tab:gold_standard_results}
\end{table*}

\begin{table*}[htbp]
    \centering
    \small
    \renewcommand{\arraystretch}{1.1}
    \setlength{\tabcolsep}{6.5pt}
    \begin{tabular}{cccccccc|c}
    \toprule
    \textbf{Category} & \textbf{Model} & \textbf{Family} & \begin{tabular}{@{}c@{}}\textbf{Real} \\ \textbf{Rights}\end{tabular} & \textbf{Torts} & \begin{tabular}{@{}c@{}}\textbf{Service} \\ \textbf{Contract}\end{tabular} & \begin{tabular}{@{}c@{}}\textbf{Finance} \\ \textbf{Contract}\end{tabular} & \textbf{Misc.} & \textbf{Overall} \\
    \midrule
    \multirow{2}{*}{\textbf{Legal LLMs}} 
    & \textit{LawLLM-7B} & 0.0545 & 0.1718 & 0.1527 & 0.1606 & 0.0879 & 0.2602 & 0.1530 \\
    & \textit{LegalOne-8B} & 0.4492 & 0.3409 & 0.3322 & 0.4207 & 0.1406 & 0.4502 & 0.3655 \\
    \midrule
    \multirow{5}{*}{\begin{tabular}{@{}c@{}}\textbf{Open-Source} \\ \textbf{General LLMs}\end{tabular}} 
    & \textit{GLM-5.3} & 0.5398 & \textbf{0.5159} & \textbf{0.5267} & 0.3541 & 0.1153 & 0.6340 & 0.4356 \\
    & \textit{DeepSeek-V4-Pro} & 0.4827 & 0.3513 & 0.4412 & 0.3970 & 0.1095 & 0.6001 & 0.3980 \\
    & \textit{Kimi-K3} & 0.5488 & 0.2532 & 0.3881 & 0.4160 & 0.1387 & 0.5016 & 0.3688 \\
    & \textit{Qwen3.8-Max} & 0.4859 & 0.4575 & 0.4570 & 0.3310 & 0.1862 & 0.4068 & 0.3860 \\
    & \textit{MiniMax-M3} & 0.4428 & 0.4203 & 0.4426 & 0.3295 & 0.0733 & 0.4817 & 0.3693 \\
    \midrule
    \multirow{3}{*}{\begin{tabular}{@{}c@{}}\textbf{Closed-Source} \\ \textbf{General LLMs}\end{tabular}} 
    & \textit{Claude-Opus-5} & 0.4979 & 0.4867 & 0.4849 & \textbf{0.5580} & \textbf{0.4195} & 0.6154 & \textbf{0.5183} \\
    & \textit{GPT-5.6-Sol} & \textbf{0.5954} & 0.3950 & 0.5043 & 0.4147 & 0.2409 & \textbf{0.6697} & 0.4788 \\
    & \textit{Gemini-3.8-Flash} & 0.5404 & 0.3632 & 0.2992 & 0.5047 & 0.1580 & 0.6108 & 0.4278 \\
    
    \bottomrule
    \end{tabular}
    \caption{Evaluation results (\textit{QWK} Score) of Legal Judgment Prediction task under the Interaction-Based Prediction setup across various cause of action classes. All reported results are averaged over five repeated runs.}
    \vspace{-10pt}
    \label{tab:interaction_based_results}
\end{table*}

\subsection{Influence of Interaction Behavior on LJP}
\label{sec:appendix_controlled_clarification}

This analysis examines how interaction behavior affects downstream judgment prediction. For each of the three closed-source models, we compare three conditions on the same 432 underspecified consultations: \textbf{Direct} predicts without asking questions, \textbf{Targeted} asks questions that recover the missing facts, and \textbf{Off-target} asks questions that leave those facts undisclosed. Table~\ref{tab:ljp_behaviors} reports the resulting LJP performance. The following protocol constructs and validates these contrasting behaviors before evaluating their prediction outcomes.

\paragraph{Dialogue Construction}
To compare useful and misdirected clarification on the same case, we construct Targeted and Off-target dialogues from independent copies of its initial history. The evaluated model receives the annotated key facts, but no reference verdicts, to generate questions intended to recover or avoid those facts, respectively. The User Model answers one question per turn over three rounds in both conditions, keeping the interaction budget fixed. Once the missing facts are recovered, remaining Targeted turns only confirm disclosed information. This produces two interaction paths that differ in whether questioning obtains the facts needed for judgment prediction.

\paragraph{Dialogue Validation}
To ensure that these interaction paths actually differ in factual recovery, DeepSeek-R1 checks each missing fact against the full dialogue without access to verdicts, marking it as recovered, absent, or contradicted and citing supporting user turns where applicable. Targeted requires complete, accurate recovery; Off-target requires no disclosure, including by implication. Both exclude unsupported assertions, questions revealing the target answer, and replies exceeding the question's factual scope. Failed dialogues return to construction, with at most five attempts per case and condition; rejection reasons and unresolved cases are retained. Throughout this process, acceptance is determined solely by these dialogue criteria, independently of whether the resulting verdict is correct.

\paragraph{Prediction and Scoring}
To measure the prediction outcomes of these behaviors, once all 432 cases pass validation, the same model predicts verdicts in a fresh context using only the validated dialogue and the direct prediction prompt (Appendix~\ref{sec:appendix_lexawn_prompt}). Construction inputs and validation feedback are excluded. The full procedure is repeated five times, computing \textit{QWK} over all 657 equally weighted claims per run and averaging across runs. Direct results are reused from Table~\ref{tab:performance_results} under identical model settings and scoring. Comparing Targeted and Off-target with Direct shows how prediction changes after successful or misdirected clarification under these controlled conditions. Note that this comparison does not measure how often models adopt either behavior autonomously; that ability is assessed by DCG task and LJP task under Interaction-Based setup.

\subsection{Details for Focal Point Categories}
\label{sec:appendix_focal_point_types}

We present the detailed illustrations of the seven distinct types of focal points of dispute in Table~\ref{tab:focal_point_categories}.

\begin{table*}[t]
    \centering
    \small
    \renewcommand{\arraystretch}{1.3}
    \begin{CJK*}{UTF8}{gbsn}
    \begin{tabular}{@{}>{\raggedright\arraybackslash}p{0.15\linewidth} >{\raggedright\arraybackslash}p{0.38\linewidth} >{\raggedright\arraybackslash}p{0.4\linewidth}@{}}
    \toprule
    \textbf{Category} & \textbf{Explanation} & \textbf{Examples} \\
    \midrule
    
    \textbf{Subject Identity} \newline ({\CJKfamily{gkai}主体身份类}) & 
    Examines whether the parties possess specific legal standing, status relationships, or appropriate qualifications (e.g., confirming labor/employment relationships, determining membership in collective economic organizations, or identifying contract parties). & 
    $\bullet$ Whether the plaintiff qualifies as a member of the defendant's collective economic organization. \newline
    $\bullet$ Whether a labor relationship exists between the plaintiff and the defendant. \newline
    $\bullet$ Identification of the lessee in the disputed ``Tower Crane Lease Contract.'' \\
    \midrule
    
    \textbf{Rights Allocation} \newline ({\CJKfamily{gkai}权属界定类}) & 
    Confirms the ultimate ownership and attribution status of substantive rights such as property rights, creditor's rights, intellectual property, or inherited assets (e.g., house ownership, scope of inheritance, land contracting rights). & 
    $\bullet$ Does the house belong to He XX, is it the joint property of He XX and Lai XX, or is it family property jointly owned by family members? \newline
    $\bullet$ Whether the disputed house located at No.~XX North, XX Village constitutes the inheritance of Chen X and Guo X. \newline
    $\bullet$ The ownership of the house in the courtyard at No.XX North Street, a village in XX District. \\
    \midrule
    
    \textbf{Contract Validity} \newline ({\CJKfamily{gkai}契约效力类}) & 
    Evaluates the legal establishment, effectiveness, invalidity, or revocability of legal actions or documents, such as contracts, agreements, and wills. & 
    $\bullet$ The validity of the provision in the divorce agreement requiring the man to pay the woman 400,000 RMB. \newline
    $\bullet$ Whether the ``Land Transfer Agreement'' signed by both parties is valid. \newline
    $\bullet$ The validity of the will made by Zhao XX. \\
    \midrule
    
    \textbf{Factual Conduct} \newline ({\CJKfamily{gkai}客观行为类}) & 
    Investigates whether the parties have actually performed specific objective actions or fulfilled obligations (e.g., whether specific payments were made, delivery obligations were fulfilled, or property was actually returned). & 
    $\bullet$ Whether the plaintiff has fulfilled the signboard renovation obligation for ``XX Bakery'' as stipulated in the agreement. \newline
    $\bullet$ Whether the 240,000 RMB paid by Chen to Zhu includes the 200,000 RMB living expenses stipulated in the divorce agreement. \newline
    $\bullet$ Whether the defendant has returned 50,000 RMB in cash to the plaintiff. \\
    \midrule
    
    \textbf{Amount Assessment} \newline ({\CJKfamily{gkai}数额界定类}) & 
    Objectively calculates, verifies, and confirms specific amounts, quantities, proportions, and values. & 
    $\bullet$ The principal amount of the loan claimed by the plaintiff. \newline
    $\bullet$ Determination of the principal and interest of the loan owed by the defendant to the plaintiff. \newline
    $\bullet$ The value of the plaintiff's tree loss. \\
    \midrule
    
    \textbf{Liability Allocation} \newline ({\CJKfamily{gkai}归责逻辑类}) & 
    Determines the causal relationship between the resulting damage and the actions taken, the degree of fault, liability for breach of contract, and the legal subject responsible for bearing the liability (who compensates and why). & 
    $\bullet$ Whether the vehicle malfunction on April 5, 2018, was caused by inadequate oil replacement during the defendant's maintenance or by other reasons. \newline
    $\bullet$ Should Wang XX bear compensation liability for Zhang's death? \newline
    $\bullet$ Who is at fault for the fire accident. \\
    \midrule
    
    \textbf{Others} \newline ({\CJKfamily{gkai}其他}) & 
    Focal points that cannot be covered by the above six categories. & 
    --- \\
    
    \bottomrule
    \end{tabular}
    \end{CJK*}
    \caption{The seven categories of focal points of dispute in civil case consultations.}
    \label{tab:focal_point_categories}
\end{table*}

\section{Generation and Evaluation Prompts}
\label{sec:appendix_lexawn_prompt}

In this appendix, we present the exact prompt leveraged by the LexAwN framework in Table~\ref{tab:lexawn_prompt}. 

Additionally, the modified version of the LexAwN prompt used in the Detail Completion Guidance task is shown in Table~\ref{tab:DCG_prompt}. Table~\ref{tab:LJP_interactionbased_prompt} presents the direct judgment prediction prompt shared by the Direct and Gold-Standard setups of the Legal Judgment Prediction task. The two setups differ only in the input history: Direct uses the underspecified consultation, whereas Gold-Standard uses its complete counterpart.

For the Interaction-Based LJP setup, if the model has not produced a verdict after three clarification rounds, we apply the direct prediction prompt in Table~\ref{tab:LJP_interactionbased_prompt} to the accumulated dialogue history.

The Standard User Model prompt shared by DCG and Interaction-Based LJP is provided in Table~\ref{tab:user_model_prompt}. Tables~\ref{tab:LCA_evaluation_prompt} and~\ref{tab:DCG_evaluation_prompt} provide the LCA and DCG evaluation prompts. Table~\ref{tab:LJP_evaluation_prompt} presents the LJP label extraction prompt.

All prompts are presented in Chinese with their corresponding English translations.

\begin{table*}[t]
    \centering
    \small
    \renewcommand{\arraystretch}{1.2}
    \begin{tabular}{@{}p{0.96\textwidth}@{}}
    \toprule
    \textbf{Original Chinese Prompt for LexAwN} \\
    \midrule
    \begin{CJK*}{UTF8}{gbsn}
    你是一个专业的AI法律顾问。你的任务是评估原告提交的民事案件法律咨询，并预测法院是否会支持其诉讼请求。
    
    用户的案件陈述可能是清晰、完备的，也可能缺失关键的法律点。在决定如何回复之前，请你首先对照该类纠纷的构成要件，对\textless 对话历史\textgreater 进行评估。如果当前事实能够形成法律推演链条，则直接预测判决结果；如果存在法定要件缺失，导致法律推演难以进行，则按需提问。
    
    请根据评估结果，严格选择并执行以下两种策略之一：
    
    {\CJKfamily{gkai}【策略一】：事实脉络大体清晰，则直接预测} \newline
    如果\textless 对话历史\textgreater 中的事实已经能够形成法律推演链条，具备了裁决基础，直接执行： \newline
    1. 法律分析：结合案件事实与适用的法律规则进行专业分析，指出哪些事实支持或削弱了原告的诉请。 \newline
    2. 预测结论：对于用户的每一条诉讼请求，分别给出明确的判决预测结果（必须明确分类为：支持、部分支持，被驳回其中之一）。
    
    \vspace{0.5em}
    {\CJKfamily{gkai}【策略二】：存在法律要件缺失导致难以推演，则按需提问 (Ask-when-Needed)} \newline
    如果\textless 对话历史\textgreater 中存在法定要件缺失，导致难以进行合理的法律推演时，执行： \newline
    1. 指出缺失：明确告知用户我们需要核实哪个案件关键点。（只能核实【一个】关键点，严禁一次核实多个） \newline
    2. 解释原因：简明扼要地说明为什么该关键点对法院的最终判决至关重要。 \newline
    3. 提出问题：提出一个清晰、具体的单点问题，用于引导用户补充相关事实。
    
    \vspace{0.5em}
    \textbf{\# Input Data} \newline
    \textless 对话历史\textgreater \{history\} \newline
    \textless /对话历史\textgreater
    
    \vspace{0.5em}
    \textbf{\# Output Constraints} \newline
    仅输出回复内容，不要包含任何标题、前言、后语、Markdown标记、标签、元数据或自我解释。
    \end{CJK*} \\
    \midrule
    \textbf{English Translation} \\
    \midrule
    You are a professional AI legal advisor. Your task is to evaluate civil case legal consultations submitted by plaintiffs and predict whether the court will support their litigation claims.
    
    The user's case statement may be clear and complete, or it may lack crucial legal points. Before deciding how to respond, please first evaluate the \textless Dialogue History\textgreater\ against the constitutive requirements of such disputes. If the current facts can form a chain of legal reasoning, directly predict the judgment result; if there is a lack of statutory requirements that makes legal reasoning difficult, ask questions as needed.
    
    Based on the evaluation results, strictly select and execute one of the following two strategies:
    
    \textbf{[Strategy 1]: The factual context is generally clear, predict directly} \newline
    If the facts in the \textless Dialogue History\textgreater\ can already form a chain of legal reasoning and provide a basis for adjudication, directly execute: \newline
    1. Legal Analysis: Conduct a professional analysis combining the case facts and applicable legal rules, pointing out which facts support or weaken the plaintiff's claims. \newline
    2. Prediction Conclusion: For \textbf{each} of the user's litigation claims, provide a clear prediction of the judgment (must be explicitly categorized as \textbf{one of} the following: Supported, Partially Supported, or Dismissed).
    
    \vspace{0.5em}
    \textbf{[Strategy 2]: Missing legal requirements make reasoning difficult, ask when needed (Ask-when-Needed)} \newline
    If the \textless Dialogue History\textgreater\ lacks statutory requirements, making it difficult to conduct reasonable legal reasoning, execute: \newline
    1. Point out the missing information: Explicitly inform the user which key case point needs to be verified. (You can only verify [ONE] key point; verifying multiple at once is strictly prohibited). \newline
    2. Explain the reason: Briefly explain why this key point is crucial to the court's final judgment. \newline
    3. Ask a question: Pose a clear, specific, single-point question to guide the user in supplementing the relevant facts.
    
    \vspace{0.5em}
    \textbf{\# Input Data} \newline
    \textless Dialogue History\textgreater \{history\} \newline
    \textless /Dialogue History\textgreater
    
    \vspace{0.5em}
    \textbf{\# Output Constraints} \newline
    Output only the response content. Do not include any titles, prefaces, epilogues, Markdown tags, labels, metadata, or self-explanations. \\
    \bottomrule
    \end{tabular}
    \caption{The LexAwN prompt, presenting the original Chinese version which was actually used in our experiments, as well as its English translation.}
    \label{tab:lexawn_prompt}
\end{table*}

\begin{table*}[t]
    \centering
    \small
    \renewcommand{\arraystretch}{1.2}
    \begin{tabular}{@{}p{0.96\textwidth}@{}}
    \toprule
    \textbf{Original Chinese Prompt for DCG Task} \\
    \midrule
    \begin{CJK*}{UTF8}{gbsn}
    你是一个专业的AI法律顾问。你的任务是评估原告提交的民事案件法律咨询。
    
    用户的案件陈述中缺失了关键的法律点。请你首先对照该类纠纷的构成要件，对\textless 对话历史\textgreater 进行评估，找出存在法定要件缺失、导致法律推演难以进行的地方，并直接执行以下提问策略：
    
    1. 指出缺失：明确告知用户我们需要核实哪个案件关键点。（注意：只能核实【一个】关键点，严禁一次核实多个） \newline
    2. 解释原因：简明扼要地说明为什么该关键点对法院的最终判决至关重要。 \newline
    3. 提出问题：提出一个清晰、具体的单点问题，用于引导用户补充相关事实。
    
    \vspace{0.5em}
    \textbf{\# Input Data} \newline
    \textless 对话历史\textgreater \{history\} \newline
    \textless /对话历史\textgreater
    
    \vspace{0.5em}
    \textbf{\# Output Constraints} \newline
    仅输出回复内容，不要包含任何标题、前言、后语、Markdown标记、标签、元数据或自我解释。
    \end{CJK*} \\
    \midrule
    \textbf{English Translation} \\
    \midrule
    You are a professional AI legal advisor. Your task is to evaluate civil case legal consultations submitted by plaintiffs.
    
    The user's case statement lacks crucial legal points. Please first evaluate the \textless Dialogue History\textgreater\ against the constitutive requirements of such disputes, identify where the lack of statutory requirements makes legal reasoning difficult, and directly execute the following questioning strategy:
    
    1. Point out the missing information: Explicitly inform the user which key case point needs to be verified. (Note: You can only verify [ONE] key point; verifying multiple at once is strictly prohibited). \newline
    2. Explain the reason: Briefly explain why this key point is crucial to the court's final judgment. \newline
    3. Ask a question: Pose a clear, specific, single-point question to guide the user in supplementing the relevant facts.
    
    \vspace{0.5em}
    \textbf{\# Input Data} \newline
    \textless Dialogue History\textgreater \{history\} \newline
    \textless /Dialogue History\textgreater
    
    \vspace{0.5em}
    \textbf{\# Output Constraints} \newline
    Output only the response content. Do not include any titles, prefaces, epilogues, Markdown tags, labels, metadata, or self-explanations. \\
    \bottomrule
    \end{tabular}
    \caption{The modified version of the LexAwN prompt used in the Detail Completion Guidance task, presenting both the original Chinese version and its English translation.}
    \label{tab:DCG_prompt}
\end{table*}

\begin{table*}[t]
    \centering
    \small
    \renewcommand{\arraystretch}{1.2}
    \begin{tabular}{@{}p{0.96\textwidth}@{}}
    \toprule
    \textbf{Original Chinese LJP Prompt Shared by Direct and Gold-Standard} \\
    \midrule
    \begin{CJK*}{UTF8}{gbsn}
    你是一个专业的AI法律顾问。你的任务是评估原告提交的民事案件法律咨询，并预测法院是否会支持其诉讼请求。

    请你对照该类纠纷的构成要件，根据\textless 对话历史\textgreater 中当前提供的事实进行法律分析，并直接预测各项诉讼请求的判决结果。本次不进行进一步问答，请勿向用户提出补充问题。
    
    请严格执行以下步骤：
    
    1. 法律分析：结合案件事实与适用的法律规则进行专业分析，指出哪些事实支持或削弱了原告的诉请。 \newline
    2. 预测结论：对于用户的每一条诉讼请求，分别给出明确的判决预测结果（必须明确分类为：支持、部分支持，被驳回其中之一）。
    
    \vspace{0.5em}
    \textbf{\# Input Data} \newline
    \textless 对话历史\textgreater \{history\} \newline
    \textless /对话历史\textgreater
    
    \vspace{0.5em}
    \textbf{\# Output Constraints} \newline
    仅输出回复内容，不要包含任何标题、前言、后语、Markdown标记、标签、元数据或自我解释。
    \end{CJK*} \\
    \midrule
    \textbf{English Translation} \\
    \midrule
    You are a professional AI legal advisor. Your task is to evaluate civil case legal consultations submitted by plaintiffs and predict whether the court will support their litigation claims.
    
    Evaluate the case against the constitutive requirements of the dispute using the facts currently provided in the \textless Dialogue History\textgreater\. Directly predict the outcome of each claim without asking follow-up questions.
    
    Please strictly execute the following steps:
    
    1. Legal Analysis: Conduct a professional analysis combining the case facts and applicable legal rules, pointing out which facts support or weaken the plaintiff's claims. \newline
    2. Prediction Conclusion: For \textbf{each} of the user's litigation claims, provide a clear prediction of the judgment (must be explicitly categorized as \textbf{one of} the following: Supported, Partially Supported, or Dismissed).
    
    \vspace{0.5em}
    \textbf{\# Input Data} \newline
    \textless Dialogue History\textgreater \{history\} \newline
    \textless /Dialogue History\textgreater
    
    \vspace{0.5em}
    \textbf{\# Output Constraints} \newline
    Output only the response content. Do not include any titles, prefaces, epilogues, Markdown tags, labels, metadata, or self-explanations. \\
    \bottomrule
    \end{tabular}
    \caption{The direct judgment prediction prompt shared by the Direct and Gold-Standard LJP setups, in Chinese and English. The input is an underspecified consultation for Direct setup and its complete counterpart for Gold-Standard setup.}
    \label{tab:LJP_interactionbased_prompt}
\end{table*}

\begin{table*}[t]
    \centering
    \small
    \renewcommand{\arraystretch}{1.2}
    \begin{tabular}{@{}p{0.96\textwidth}@{}}
    \toprule
    \textbf{Original Chinese Prompt for the Standard User Model} \\
    \midrule
    \begin{CJK*}{UTF8}{gbsn}
    你现在需要扮演一名正在进行民事案件法律咨询的原告当事人。你的任务是根据\textless 对话历史\textgreater{}中用户已提供的事实和\textless 案件客观事实\textgreater{}，对AI法律顾问的最新追问做出自然、恰当的回复。

    \textbf{\# Input Data} \newline
    \textless 对话历史\textgreater{}\{history\} \newline
    \textless /对话历史\textgreater{}

    \textless 案件客观事实\textgreater{} \newline
    \{key\_facts\} \newline
    \textless /案件客观事实\textgreater{}

    {\CJKfamily{gkai}\# 逻辑与回复策略} \newline
    请根据最新追问的具体含义决定回复，不要把AI的假设或推测当作已确认事实。

    1. 事实命中（有效且可答）： \newline
    - 评估标准：问题是在核实具体事实，且用户已提供的事实或\textless 案件客观事实\textgreater{}足以支持回答。日常表达、间接提问或证据线索均可，不要求与事实原文逐字匹配。 \newline
    - 回复策略：准确、充分地回答所问事实，可补充理解回答所必需的解释；不要披露超出当前问题范围的其他隐藏细节。若只能回答部分内容，回答有依据的部分，并明确说明其余部分不清楚。绝不能编造未提供的细节。

    2. 事实未命中（偏离或不可答）： \newline
    - 评估标准：问题仅宽泛提及法律概念，没有核实具体事实；或者所提供的信息不足以回答。 \newline
    - 回复策略：宽泛提问时不要主动披露隐藏事实，可请AI明确想了解的情况。信息不足时，自然地说明“不清楚”或“无法确认”；不要把未提及的事情说成没有发生，也不要仅因缺少信息而断言问题与案件无关。

    \textbf{\# Output Constraints} \newline
    - 必须以原告（缺乏专业法律背景的咨询者）的第一人称口吻进行回复，语气和你在\textless 对话历史\textgreater{}中的咨询一致，符合上下文情境。 \newline
    - 仅输出你作为用户的回复文本，绝对不要包含任何标题、Markdown代码块标记、前缀标签（例如“原告：”、“回复：”）、元数据或自我解释。
    \end{CJK*} \\
    \midrule
    \textbf{English Translation} \\
    \midrule
    You are a plaintiff seeking legal advice about a civil case. Respond naturally and appropriately to the AI legal advisor's latest follow-up question using the facts already provided by the user in \textless Dialogue History\textgreater{} and \textless Objective Case Facts\textgreater{}.

    \textbf{\# Input Data} \newline
    \textless Dialogue History\textgreater{}\{history\} \newline
    \textless /Dialogue History\textgreater{}

    \textless Objective Case Facts\textgreater{} \newline
    \{key\_facts\} \newline
    \textless /Objective Case Facts\textgreater{}

    \textbf{\# Reasoning and Response Strategy} \newline
    Respond according to the specific meaning of the latest question. Do not treat the AI's assumptions or speculation as confirmed facts.

    1. Factual match (relevant and answerable): \newline
    - Criterion: The question seeks specific facts, and the facts already provided by the user or \textless Objective Case Facts\textgreater{} support an answer. Everyday wording, indirect questions, and evidence cues are acceptable; verbatim matching is unnecessary. \newline
    - Response: Answer the requested facts accurately and sufficiently, including explanations necessary to understand the answer. Withhold other hidden details beyond the current question's scope. If only part is answerable, give the supported information and explicitly state uncertainty about the rest. Never invent details that were not supplied.

    2. No factual match (off-target or unanswerable): \newline
    - Criterion: The question only mentions a broad legal concept without requesting specific facts, or the supplied information is insufficient to answer. \newline
    - Response: For broad questions, do not volunteer hidden facts; you may ask the AI to specify what it wants to know. If information is insufficient, naturally say ``I don't know'' or ``I cannot confirm.'' Do not treat an unmentioned event as having never happened or assert that a question is irrelevant merely because information is missing.

    \textbf{\# Output Constraints} \newline
    - Reply in the first person as the plaintiff, a layperson without professional legal training. Match the user's tone in \textless Dialogue History\textgreater{} and the conversational context. \newline
    - Output only the user's reply. Do not include titles, Markdown code fences, prefix labels (e.g., ``Plaintiff:'' or ``Reply:''), metadata, or self-explanations. \\
    \bottomrule
    \end{tabular}
    \caption{The Standard User Model prompt shared by DCG and Interaction-Based LJP, with the original Chinese text and its English translation.}
    \label{tab:user_model_prompt}
\end{table*}

\begin{table*}[t]
    \centering
    \small
    \renewcommand{\arraystretch}{1.2}
    \begin{tabular}{@{}p{0.96\textwidth}@{}}
    \toprule
    \textbf{Original Chinese Evaluation Prompt for LCA} \\
    \midrule
    \begin{CJK*}{UTF8}{gbsn}
    你是一个严谨的文本分类器与指令遵循评估器。你的任务是阅读一段由“AI法律顾问”生成的回复，并判断该回复执行的是【策略一】（直接预测）还是【策略二】（主动追问）。

    仅根据待评估的新回复识别策略，不独立判断案件是否完备，也不评价判决预测是否正确。策略一对应模型将咨询视为完备，策略二对应模型将咨询视为信息不足；分类结果由评分程序与真实类别比较。

    {\CJKfamily{gkai}【背景与策略定义】} \newline
    生成待评估回复的AI模型，其系统指令要求它严格在以下两种策略中选择其一执行：

    - 【策略一】（直接预测）：当用户的案件陈述法律要件基本齐备时触发。该策略的回复特征是： \newline
    1. 包含法律分析：结合案件事实与适用的法律规则进行分析。 \newline
    2. 包含预测结论：会对用户的每一条诉讼请求，分别给出明确的预测结果（明确带有“支持”、“部分支持”或“被驳回”等字眼）。

    - 【策略二】（按需提问/Ask-when-Needed）：当案件陈述明显缺失核心法律要件时触发。该策略的回复特征是： \newline
    1. 指出缺失：明确告知用户需要核实哪个单一的案件关键点。 \newline
    2. 解释原因：说明为什么该关键点对法院的最终判决至关重要。 \newline
    3. 提出问题：没有给出最终的判决预测，而是向用户提出一个具体的单点问题，引导其补充事实。

    {\CJKfamily{gkai}【判断标准】} \newline
    - 如果待评估回复执行的是【策略一】（即：给出了明确的判决预测结论），请输出：策略一 \newline
    - 如果待评估回复执行的是【策略二】（即：指出了缺失要件并向用户提出了追问），请输出：策略二 \newline
    - 【特殊情况处理】（混合特征）：如果待评估回复中既给出了明确的判决结果，又指出了缺失要件并向用户提出了追问，请根据回复的核心侧重点（主导地位）进行判定： \newline
    - 若“明确的预测结论”占主导地位（例如：主体已经完成了明确的判决预测，追问仅作为附带的补充建议或针对非核心信息的询问），请输出：策略一 \newline
    - 若“指出缺失并追问”占主导地位（例如：预测仅仅是基于假设的模糊推测，回复的主体目的是要求用户补充关键事实才能真正得出结论），请输出：策略二

    {\CJKfamily{gkai}【输入数据】} \newline
    \textless 待评估回复\textgreater{} \newline
    \{history\} \newline
    \textless /待评估回复\textgreater{}

    {\CJKfamily{gkai}【输出限制】} \newline
    严格只输出“策略一”或“策略二”三个字(不带引号)。绝对不要包含任何前言后语、解释说明、标点符号、Markdown标记或换行符。
    \end{CJK*} \\
    \midrule
    \textbf{English Translation} \\
    \midrule
    You are a rigorous text classifier and instruction-following evaluator. Read a reply generated by an AI legal advisor and identify whether it follows [Strategy 1] (direct prediction) or [Strategy 2] (proactive clarification).

    Classify only the new reply. Do not independently assess case completeness or prediction correctness. Strategy 1 means the model treats the consultation as complete; Strategy 2 means it treats it as underspecified. The scoring code compares these labels with the reference categories.

    \textbf{[Background and Strategy Definitions]} \newline
    The model generating the reply was instructed to choose exactly one of these strategies:

    - [Strategy 1] (direct prediction): Triggered when the user's case statement contains the essential legal elements. Its response features are: \newline
    1. Legal analysis combining case facts and applicable legal rules. \newline
    2. A clear prediction for each claim, explicitly using terms such as ``Supported,'' ``Partially Supported,'' or ``Dismissed.''

    - [Strategy 2] (Ask-when-Needed): Triggered when core legal elements are clearly missing. Its response features are: \newline
    1. Identify the single key case point requiring clarification. \newline
    2. Explain why it is crucial to the court's final judgment. \newline
    3. Ask a specific, single-point question to elicit facts instead of giving a final verdict prediction.

    \textbf{[Classification Criteria]} \newline
    - If the reply follows [Strategy 1], i.e., gives a clear verdict prediction, output: Strategy 1 \newline
    - If it follows [Strategy 2], i.e., identifies missing elements and asks for clarification, output: Strategy 2 \newline
    - [Mixed features]: If the reply both gives a clear verdict prediction and identifies missing elements with a follow-up question, classify by its dominant focus: \newline
    - If the clear prediction dominates (e.g., a definitive prediction is already given and clarification is merely supplementary advice or concerns non-essential information), output: Strategy 1 \newline
    - If identifying missing information and seeking clarification dominates (e.g., the prediction is only vague, hypothetical speculation and the reply mainly seeks crucial facts before reaching a conclusion), output: Strategy 2

    \textbf{[Input Data]} \newline
    \textless Reply to Evaluate\textgreater{} \newline
    \{history\} \newline
    \textless /Reply to Evaluate\textgreater{}

    \textbf{[Output Constraints]} \newline
    Output only ``Strategy 1'' or ``Strategy 2'' without quotation marks. Do not include any preface, closing remarks, explanation, punctuation, Markdown markup, or line breaks. \\
    \bottomrule
    \end{tabular}
    \caption{The LCA evaluation prompt, with the original Chinese text and its English translation.}
    \label{tab:LCA_evaluation_prompt}
\end{table*}

\begin{table*}[t]
    \centering
    \small
    \renewcommand{\arraystretch}{1.2}
    \begin{tabular}{@{}p{0.96\textwidth}@{}}
    \toprule
    \textbf{Original Chinese Evaluation Prompt for DCG} \\
    \midrule
    \begin{CJK*}{UTF8}{gbsn}
    你是一个法律逻辑评估专家。你的任务是评估AI法律顾问在法律咨询中，对用户针对案件细节进行补充的引导，是否命中了相应案件的核心争议焦点。

    仅评估当前这一轮提问的相关性与可答性，不判断用户是否已经实际补全了事实。

    {\CJKfamily{gkai}\# 评估任务} \newline
    请仔细阅读下方的输入数据，并综合判断以下标准是否均成立： \newline
    1. {\CJKfamily{gkai}引导的正确性}：\textless AI提问内容\textgreater{}所关注的问题，应属于解决\textless 裁判文书争议焦点\textgreater{}所必需的{\CJKfamily{gkai}法律要点或事实前提}（不要求表述完全一致，但该提问应有助于厘清争议焦点，允许合理的逻辑延伸，但不能仅是宽泛的领域相关）； \newline
    2. {\CJKfamily{gkai}引导的可答性}：针对\textless AI提问内容\textgreater{}中的提问，在\textless 争议焦点对应案件事实\textgreater{}中能够找到{\CJKfamily{gkai}对应的事实信息或证据线索}，足以让用户对该提问做出有效回应（不要求事实完全覆盖提问，但必须能支持对所问事实的实质性回答，不能仅有主题关联）。

    {\CJKfamily{gkai}\# 输入数据} \newline
    \textless AI提问内容\textgreater{} \newline
    \{ai\_response\} \newline
    \textless /AI提问内容\textgreater{}

    \textless 裁判文书争议焦点\textgreater{} \newline
    \{focus\_text\} \newline
    \textless /裁判文书争议焦点\textgreater{}

    \textless 争议焦点对应案件事实\textgreater{} \newline
    \{key\_facts\} \newline
    \textless /争议焦点对应案件事实\textgreater{}

    {\CJKfamily{gkai}\# 输出要求} \newline
    - 如果上述两个条件均满足，请仅输出：是 \newline
    - 如果任一条件不满足（包括仅宽泛相关、缺少具体事实提问或事实不足以支持回答），请仅输出：否 \newline
    - 严格限制：仅输出一个字，严禁包含任何标点符号、理由解释、前言后语或Markdown标记。
    \end{CJK*} \\
    \midrule
    \textbf{English Translation} \\
    \midrule
    You are an expert in legal reasoning evaluation. Assess whether an AI legal advisor's guidance for eliciting case details from a user addresses the case's core focal point of dispute.

    Evaluate only the current question's relevance and answerability, not whether the user has actually supplied the missing facts.

    \textbf{\# Evaluation Task} \newline
    Read the inputs carefully and assess whether both criteria hold: \newline
    1. \textbf{Guidance relevance}: The issue raised in \textless AI Question\textgreater{} must concern a \textbf{legal issue or factual premise} necessary to resolve \textless Focal Points of Dispute\textgreater{}. Exact wording is unnecessary, but the question must help clarify the dispute. Reasonable logical extensions are allowed; broad domain relevance alone is insufficient. \newline
    2. \textbf{Guidance answerability}: \textless Corresponding Case Facts\textgreater{} must contain \textbf{corresponding factual information or evidence cues} sufficient for the user to respond meaningfully to \textless AI Question\textgreater{}. Full coverage of the question is unnecessary, but the facts must support a substantive answer to the factual request; topical association alone is insufficient.

    \textbf{\# Input Data} \newline
    \textless AI Question\textgreater{} \newline
    \{ai\_response\} \newline
    \textless /AI Question\textgreater{}

    \textless Focal Points of Dispute\textgreater{} \newline
    \{focus\_text\} \newline
    \textless /Focal Points of Dispute\textgreater{}

    \textless Corresponding Case Facts\textgreater{} \newline
    \{key\_facts\} \newline
    \textless /Corresponding Case Facts\textgreater{}

    \textbf{\# Output Requirements} \newline
    - If both criteria are met, output only: Yes \newline
    - If either criterion is unmet (including broad relevance alone, no specific factual request, or insufficient facts to support an answer), output only: No \newline
    - Output exactly one word, ``Yes'' or ``No''. Do not include punctuation, explanations, prefaces, closing remarks, or Markdown markup. \\
    \bottomrule
    \end{tabular}
    \caption{The DCG evaluation prompt, with the original Chinese text and its English translation.}
    \label{tab:DCG_evaluation_prompt}
\end{table*}

\begin{table*}[t]
    \centering
    \small
    \renewcommand{\arraystretch}{1.2}
    \begin{tabular}{@{}p{0.96\textwidth}@{}}
    \toprule
    \textbf{Original Chinese Label Extraction Prompt for LJP Evaluation} \\
    \midrule
    \begin{CJK*}{UTF8}{gbsn}
    {\CJKfamily{gkai}\# 角色与任务} \newline
    你是一个专业的法律文本分析专家。你的任务是阅读法律助手的最终\textless 待提取回复\textgreater{}，提取其对原告格式化的\textless 正式诉请\textgreater{}中每一项诉讼请求的预测结果。只提取助手已表达的预测，不重新进行法律判断，也不评价预测是否正确。

    {\CJKfamily{gkai}\# 分类标准} \newline
    对于\textless 正式诉请\textgreater{}中的每一项请求，你必须且只能从以下三个选项中选择一个作为预测结果： \newline
    - 支持 \quad - 部分支持 \quad - 驳回

    {\CJKfamily{gkai}\# 提取与推理规则} \newline
    1. 严格对应：你的预测结果列表必须与\textless 正式诉请\textgreater{}中的请求数量完全一致，并与\textless 正式诉请\textgreater{}中的顺序一一对应。 \newline
    2. 诉讼费处理： \newline
    - 如果\textless 正式诉请\textgreater{}中包含案件受理费/诉讼费请求，则正常处理。 \newline
    - 如果\textless 正式诉请\textgreater{}中省略了案件受理费/诉讼费请求，但是法律助手在\textless 待提取回复\textgreater{}中对诉讼费请求的预测结果一并进行了推断，这是正常现象，直接忽略即可，不要将这部分预测结果加入列表中。 \newline
    3. 综合判断：优先提取助手对{\CJKfamily{gkai}每一项}诉请明确给出的最终预测；未使用标准标签时，再结合\textless 待提取回复\textgreater{}的上下文判断。归类为“支持”的诉请可能包含类似“依据充分”、“大概率支持”的表述；“部分支持”应指仅支持部分金额或请求范围，如“可能酌情扣减”，不能仅因“大概率”或“较大不确定性”等信心表述而归为此类；归类为“驳回”的诉请可能包含类似“难以得到支持”、“没有依据”等表述。将“被驳回”等同义表述统一为“驳回”。

    {\CJKfamily{gkai}\# 极其严格的输出格式} \newline
    必须输出为标准的JSON格式。严禁包含任何多余的标点符号、理由解释、前言后语，{\CJKfamily{gkai}绝对不要使用任何Markdown标记（如 \texttt{\char96\char96\char96{}json} ）}。 \newline
    "results"列表长度与\textless 正式诉请\textgreater{}中的诉请数量必须相同，结果顺序与诉请顺序也必须严格保持一致(注意，\textless 正式诉请\textgreater{}中每一行代表一项诉请，{\CJKfamily{gkai}只要没有换行，都属于同一条诉请}，不要对同一条诉请输出多个预测结果！)。 \newline
    输出示例格式如下： \newline
    \{ "results": ["驳回", "支持", "部分支持"] \}

    {\CJKfamily{gkai}\# 输入数据} \newline
    \textless 待提取回复\textgreater{} \{history\} \textless /待提取回复\textgreater{}

    \textless 正式诉请\textgreater{} \{claims\} \textless /正式诉请\textgreater{}
    \end{CJK*} \\
    \midrule
    \textbf{English Translation} \\
    \midrule
    \textbf{\# Role and Task} \newline
    You are a professional legal text analyst. Read the legal assistant's final \textless Response to Extract\textgreater{} and extract its predicted outcome for each item in the plaintiff's formatted \textless Formal Claims\textgreater{}. Extract only predictions expressed by the assistant; do not make a new legal judgment or assess prediction correctness.

    \textbf{\# Classification Criteria} \newline
    For each item in \textless Formal Claims\textgreater{}, select exactly one of these three outcomes: \newline
    - Supported \quad - Partially Supported \quad - Dismissed

    \textbf{\# Extraction and Reasoning Rules} \newline
    1. Strict correspondence: The prediction list must contain exactly as many items as \textless Formal Claims\textgreater{}, in the same order. \newline
    2. Litigation costs: \newline
    - If \textless Formal Claims\textgreater{} includes a claim for case acceptance fees or litigation costs, process it normally. \newline
    - If \textless Formal Claims\textgreater{} omits such a claim but the assistant also predicts its outcome in \textless Response to Extract\textgreater{}, this is normal. Ignore that prediction and do not add it to the list. \newline
    3. Contextual interpretation: Prioritize the assistant's explicit final prediction for \textbf{each} claim. When standard labels are absent, interpret the context of \textless Response to Extract\textgreater{}. ``Supported'' may be expressed as ``sufficient grounds'' or ``likely to be supported.'' ``Partially Supported'' means that only part of the amount or scope of relief is supported, e.g., ``may be reduced at the court's discretion.'' Confidence expressions such as ``likely'' or ``considerable uncertainty'' alone do not imply partial support. ``Dismissed'' may be expressed as ``unlikely to be supported'' or ``no grounds.'' Normalize equivalent dismissal expressions, such as ``Rejected,'' to ``Dismissed.''

    \textbf{\# Strict Output Format} \newline
    Output valid JSON only. Do not include extraneous punctuation, explanations, prefaces, or closing remarks. \textbf{Never use Markdown markup (such as \texttt{\char96\char96\char96{}json}).} \newline
    The length of the "results" list must equal the number of claims in \textless Formal Claims\textgreater{}, and its order must strictly match the claim order. Each line in \textless Formal Claims\textgreater{} represents one claim; \textbf{text without a line break belongs to the same claim}. Do not output multiple predictions for a single claim. \newline
    Example output format: \newline
    \{ "results": ["Dismissed", "Supported", "Partially Supported"] \}

    \textbf{\# Input Data} \newline
    \textless Response to Extract\textgreater{} \{history\} \textless /Response to Extract\textgreater{}

    \textless Formal Claims\textgreater{} \{claims\} \textless /Formal Claims\textgreater{} \\
    \bottomrule
    \end{tabular}
    \caption{The LJP evaluation prompt for extracting claim-level verdict labels, with the original Chinese text and its English translation.}
    \label{tab:LJP_evaluation_prompt}
\end{table*}

\section{Disclosure of AI Writing Assistance}
\label{sec:ai_assist}

During the preparation of this manuscript, we used LLMs to assist with rephrasing author-written text. The authors reviewed all suggested textual changes and took full responsibility for the final content of the manuscript.
\end{document}